\documentclass{article}

\usepackage[preprint]{neurips_2024}

\usepackage{listings}

\usepackage{alltt}
\usepackage[utf8]{inputenc} % allow utf-8 input
\usepackage[T1]{fontenc}    % use 8-bit T1 fonts
\usepackage{hyperref}       % hyperlinks
\usepackage{url}            % simple URL typesetting
\usepackage{booktabs}       % professional-quality tables
\usepackage{amsfonts}       % blackboard math symbols
\usepackage{nicefrac}       % compact symbols for 1/2, etc.
\usepackage{microtype}      % microtypography
\usepackage{xcolor}         % colors
\usepackage{amsmath}
\usepackage{graphicx}
\usepackage{mathrsfs}
\usepackage{subcaption}
\usepackage{wrapfig}

\title{\textsc{YouDream} : Generating Anatomically Controllable Consistent Text-to-3D Animals}

\author{
Sandeep Mishra$^\dagger$\thanks{Equal contribution.}~~,
Oindrila Saha$^\ddagger$\footnote[1]~~~,
and Alan C. Bovik$^\dagger$\\
$^\dagger$University of Texas at Austin, 
$^\ddagger$University of Massachusetts Amherst\\
\texttt{sandy.mishra@utexas.edu}, \texttt{osaha@umass.edu}, \texttt{bovik@ece.utexas.edu}
}

\begin{document}

\maketitle

\begin{abstract}
3D generation guided by text-to-image diffusion models enables the creation of visually compelling assets. However previous methods explore generation based on image or text. The boundaries of creativity are limited by what can be expressed through words or the images that can be sourced. We present \textsc{YouDream}, a method to generate high-quality anatomically controllable animals. \textsc{YouDream} is guided using a text-to-image diffusion model controlled by 2D views of a 3D pose prior. Our method generates 3D animals that are not possible to create using previous text-to-3D generative methods. Additionally, our method is capable of preserving anatomic consistency in the generated animals, an area where prior text-to-3D approaches often struggle. Moreover, we design a fully automated pipeline for generating commonly found animals. To circumvent the need for human intervention to create a 3D pose, we propose a multi-agent LLM that adapts poses from a limited library of animal 3D poses to represent the desired animal. A user study conducted on the outcomes of \textsc{YouDream} demonstrates the preference of the animal models generated by our method over others. Turntable results and code are released at this \href{https://youdream3d.github.io/}{URL}.

\end{abstract}

\begin{figure}
    \centering
    \includegraphics[width=1.0\linewidth]{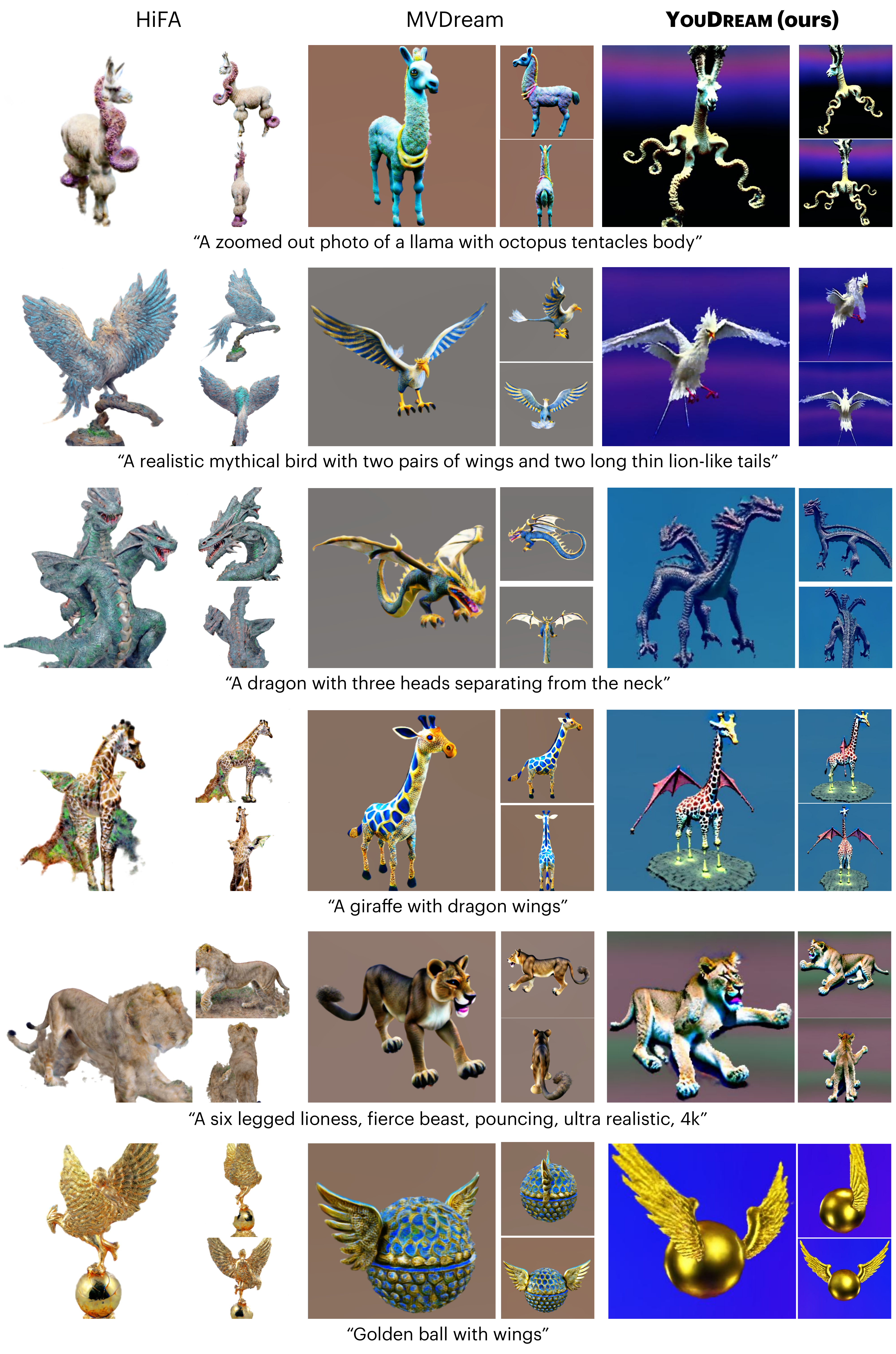}
    \caption{\textbf{Creating unreal creatures.} Our method generates imaginary creatures based on an artist's creative control. We show that these creatures cannot be generated faithfully only based on text. Each row depicts a 3D animal generated by HiFA, MVDream, and \textsc{YouDream} (left to right) using the prompt mentioned below the row. We present 3D pose controls used to create these in the Sec.~\ref{sec:implementation_details_suppl} (results best viewed zoomed in).}
    \label{fig:creative_creation}
\end{figure}

\section{Introduction}
Text-to-3D generative modeling using diffusion models has seen fast-paced growth recently with methods utilizing text-to-image (T2I)~\cite{poole2022dreamfusion, chen2023fantasia3d, zhu2023hifa, seo2023let}, (text+camera)-to-image (TC2I)~\cite{shi2023mvdream, li2023mvcontrol} and (image+camera)-to-image (IC2I)~\cite{liu2023zero, wang2023imagedream, ye2023consistent} diffusion models. These methods are widely accepted by AI enthusiasts, content creators, and 3D artists to create high-quality 3D content. However, generating 3D assets using such methods is dependent on what can be expressed through text or the availability of an image faithful to the user's imagination. In this work, we provide more control to the artist to bring their creative imagination to life. \textsc{YouDream} can generate high-quality 3D animals based on any 3D skeleton, by utilizing a 2D pose-controlled diffusion model which generates images adhering to 2D views of a 3D pose. Using depth, edge, and scribble has also been explored for controllable image generation~\cite{zhang2023adding}. However, in a 3D context, pose offers both 3D consistency as well as room for creativity. Other controls are restrictive as edge/depth/boundary of 2D views of a pre-existing object needs to be used to provide control, which limits the generated shape to be very similar to the existing asset. We show that the multi-view consistency offered by our 3D pose prior results in the generation of anatomically and geometrically consistent animals. Creating 3D pose control also requires minimal human effort. To further alleviate this effort, we also present a multi-agent LLM setup that generates 3D poses for novel animals commonly observed in nature.

3D generation guided by T2I models involves using the gradient computed by Score Distillation Sampling (SDS)~\cite{poole2022dreamfusion} to optimize a 3D representation such as a NeRF~\cite{mildenhall2021nerf}. During any intermediate step of the training, a rendered image captured by a random camera is added to Gaussian noise and passed to a T2I diffusion model, along with a directional prompt. The diffusion model estimates the added noise, which in turn is used to create a denoised image. In effect, this process pushes the rendered image of the NeRF representation slightly closer to the denoised image during each iteration. Thus, any unwanted semantic or perceptual issues arising in the denoised image are also transferred to the NeRF. This is especially problematic for deformable objects such as animals, where variations in pose over views often results in the Janus-head problem, dehydrated assets, and geometric and anatomical inconsistencies. 

% To tackle these issues we present a method to generate high quality, anatomically consistent and controllable animals utilizing a text-to-image TetraPose ControlNet. \textsc{YouDream} acheives multi-view consistency by strictly adhering to a 3D pose control.

TC2I diffusion models, which encode camera parameters and train using 3D objects from various views learn multi-view consistency and thus are able to produce better geometries. However, they lack in diversity owing to the limited variation in training data, as compared to text-to-image models. Along with this, methods using IC2I diffusion models also face the problems arising from Novel View Synthesis (NVS), which requires hallucination of unseen regions along with accurate geometric transformation of observed parts. While these camera guided diffusion models perform better than T2I models in many cases, their limited diversity and lack of control limit the creativity of their users. By utilizing a 3D pose prior, \textsc{YouDream} consistently outperforms previous methods that use T2I diffusion models, in terms of generating biologically plausible animals. Despite not being trained on any 3D data, our method also outperforms the 3D-aware TC2I diffusion model MVDream~\cite{shi2023mvdream} in text-to-3D animal generation in terms of ``Naturalness'', ``Text-Image Alignment'' and CLIP score (see Sec.~\ref{sec:experiments}).

3D-consistency for human avatar creation has been explored extensively in recent works~\cite{cao2023dreamavatar, huang2024dreamwaltz, kolotouros2024dreamhuman, hong2022avatarclip, zhang2024avatarverse, zhang2022avatargen}. These models rely on a 3D human pose and shape prior, usually the SMPL~\cite{loper2023smpl} or SMPL-X~\cite{SMPL-X:2019} model. This strategy can represent a variety of geometrically consistent human avatars. However, representing the animal kingdom is challenging owing to its immense diversity which cannot be represented using any existing parametric models. Sizes and shapes vary considerably across birds, reptiles, mammals, and amphibians, hence until now, no single shape or pose prior exists that can represent all tetrapods. Parametric models such as SMAL~\cite{Zuffi:CVPR:2017} and MagicPony~\cite{wu2023magicpony} suffer from severe diversity issues, and hence cannot be used as pose or shape prior. Thus to circumvent human effort in generating a 3D pose prior for animals prevalent in nature, we present a method for automatic generation of diverse 3D poses using a multi-agent LLM supported by a small library of animal 3D poses. Additionally, we present a method to automatically generate an initial shape based on a 3D pose, which is utilized for NeRF initialization. Code for the whole framework will be open-sourced upon acceptance.

In summary, \textsc{YouDream} offers the following key contributions:

\begin{itemize}
    \item a TetraPose ControlNet, trained on tetrapod animals across various families, that enables the generation of diverse animals at test time, both real and unreal.
    % , following a single format of pose control. 
    \item a multi-agent LLM that can generate the 3D pose of any desired animal in a described state, supported by a small library of 16 predefined animal 3D poses for reference.  
    \item a user-friendly tool to create/modify 3D poses for unreal creatures. The same tool automatically generates an initial shape based on the 3D skeleton.
    \item a pipeline to generate geometrically and anatomically consistent animals based on an input text by adhering to a 3D pose prior.
\end{itemize}

% Talk about image-to-3d

% discuss about 3dfuse, how depth is not good

\section{Related Work}
% Generating 3D assets without specific conditions requires understanding the diverse distributions of 3D data. There are two primary strategies: explicit and implicit. Structured representations such as point clouds~\cite{achlioptas2018learning, luo2021diffusion}, voxel grids~\cite{lin2023infinicity, smith2017improved}, and mesh models~\cite{zhang2021sketch2model} are categorized as explicit methods. Implicit techniques generally rely on abstract representations, such as signed distance functions (SDFs)~\cite{chen2019learning, cheng2023sdfusion, mittal2022autosdf}, tri-planes~\cite{chen2023single}, the parameters of multi-layer perceptrons (MLPs)~\cite{erkocc2023hyperdiffusion}, and radiance fields~\cite{lorraine2023att3d}. 

The field of \textbf{3D animal generation} has rapidly advanced due to studies that offer methods and insights for modeling animal structures and movements in 3D. SMAL~\cite{Zuffi:CVPR:2017} introduced a method to fit a parametric 3D shape model, derived from 3D scans, to animal images using 2D keypoints and segmentation masks, with extensions to multi-view images~\cite{zuffi2018lions}. The variety of animals able to be represented by SMAL is severely limited. Subsequent efforts, such as LASSIE~\cite{yao2022lassie, yao2023hi, yao2024artic3d}, have focused on deriving 3D shapes directly from smaller image collections by identifying self-supervised semantic correspondences to discover 3D parts. Succeeding work represent animals using a parametric model~\cite{jakab2023farm3d, wu2023dove, wu2023magicpony, li2024learning} learnt from images or videos. Despite these advances, these methods are class-specific and lack in the diversity of animals that can be represented. \textsc{YouDream}is able to generate a great variety of animals including those that have not been observed previously with higher details (Fig.~\ref{fig:vs3dfauna}).

High quality \textbf{text-to-3D asset generation} has been fueled by the availability of large-scale diverse datasets of text-image pairs and the success of text-to-image contrastive and generative models trained on them. Contrastive methods such as CLIP~\cite{radford2021learning} and ALIGN~\cite{jia2021scaling} learn a common embedding between visual and natural language domains. Generative methods like Imagen~\cite{saharia2022photorealistic} and Stable Diffusion~\cite{rombach2022high} utilize a diffusion model to learn to generate images given text latents. These methods inherently learn to understand the appearance of entities across various views and poses. Text-to-3D generative modeling methods~\cite{mohammad2022clip, jain2022zero, wang2023score, poole2022dreamfusion} exploit this information by using these text-image models to guide the creation of 3D representations by NeRFs~\cite{mildenhall2021nerf}. The quality of 3D assets produced by these early methods suffer from several issues such as smooth geometries, saturated appearances, as well as geometric issues such as the Janus (multi-head) problem. Subsequent recent methods have ameliorated these problems by the use of modified loss functions~\cite{wang2024prolificdreamer, zhu2023hifa}, using Deep Marching Tetrahedra~\cite{shen2021deep} for 3D representation~\cite{chen2023fantasia3d}, and modified negative prompt weighing strategies~\cite{armandpour2023re}. However these methods still fail to produce anatomically correct animals, often producing implausible geometries or even extra or insufficient limbs. Prior work 3DFuse~\cite{seo2023let} uses sparse point clouds predicted from images as depth control for T2I diffusion, however still produces anatomically inconsistent animals due to the inaccuracy of image-to-point cloud predictors and a high dependency on the initial generated image (Fig.~\ref{fig:common_animal}). Recently, 3D-aware diffusion models trained on paired text-3D datasets by encoding camera parameters have been used to generate 3D assets~\cite{Tang2023mvdiffusion, shi2023mvdream}. As these methods learn using various views of 3D objects, they rarely produce geometric inconsistency. However these methods are limited by the variety of 3D data available, which is quite scarce as compared to image data that T2I diffusion models have been trained on. These are trained using 3D object databases such as Objaverse~\cite{deitke2023objaverse} and Objaverse-XL~\cite{deitke2024objaverse} which are considerably smaller than text-image paired datasets such as LAION-5B~\cite{schuhmann2022laion} used for training T2I diffusion models. Thus, they often struggle to follow the text input faithfully in case of complex prompts (Fig.~\ref{fig:creative_creation}). By comparison, our method accurately follows the text prompt owing to the use of T2I diffusion models trained on vast image data. \textsc{YouDream} strictly adheres to input 3D pose priors, thus producing geometrically consistent and anatomically correct animals. 

\textbf{Large Language Models (LLMs)} have been explored in the context of \textbf{3D generation and editing} previously. LLMs have been used~\cite{yin2023shapegpt, siddiqui2023meshgpt} to generate and edit shapes using an embedding space trained on datasets such as ShapeNet~\cite{chang2015shapenet}. Prior work have also used LLMs to generate code for 3D modeling software, such as Blender, to create objects~\cite{yuan20243d} and scenes~\cite{sun20233d, hu2024scenecraft}. These methods produce impressive results suggesting at LLMs' 3D understanding capability, but explore limited variety of generation often limited to shapes, or generate layouts/scenes. 3D pose generation with LLMs using text as input has been recently explored for humans. ChatPose~\cite{feng2024chatpose} and MotionGPT~\cite{zhang2024motiongpt} generate pose parameters for a SMPL model based on textual input. LLMs have also been previously shown to accurately reason about anatomical differences of animals~\cite{menon2022visual, saha2024improved}. In this work, we show a novel application of off-the-shelf LLMs for generalized 3D pose generation based on the name of an animal supported by a library of animal 3D poses.

\textbf{User-controlled generation} has been introduced in several studies~\cite{zhang2023adding, mou2024t2i}, and has gained widespread adoption among artists for crafting remarkable illustrations, including artistic QR codes to interior designs. However, the use of user control in 3D is still under-explored. Recent works such as MVControl~\cite{li2023mvcontrol} and Control3D~\cite{chen2023control3d} guide the 3D generation process using a 2D condition image of a single view. By contrast, the generation process in \textsc{YouDream} is guided using 2D views of a 3D pose which is dependent on the sampled camera pose. This strategy not only allows \textsc{YouDream} to take in specialized user control but also ensures multi-view geometric consistency.

% These innovations have paved the way for the utilization of the knowledge of large pre-trained T2I models for 3D asset generation, primarily those that are CLIP-guided, or are 2D diffusion-guided. CLIP-guided methods such as~\cite{mohammad2022clip, jain2022zero} leverage cross-modal matching models like CLIP~\cite{radford2021learning} for text-to-3D conversion, whereas diffusion-guided strategies utilize text-to-image diffusion models~\cite{zhu2023hifa, chen2023fantasia3d}, like Imagen~\cite{saharia2022photorealistic} and Stable Diffusion~\cite{rombach2022high}, to generate 3D assets from textual descriptions. T2I diffusion models have been noted to deliver superior text-to-3D generation performance, employing techniques such as Score Distillation Sampling (SDS)~\cite{poole2022dreamfusion}, which refines noise in images captured from NeRFs~\cite{mildenhall2021nerf}, and Score-Jacobian-Chaining~\cite{wang2023score}, which aggregates image gradients into 3D asset gradients. 

\begin{figure}
    \centering
    \includegraphics[width=1\linewidth]{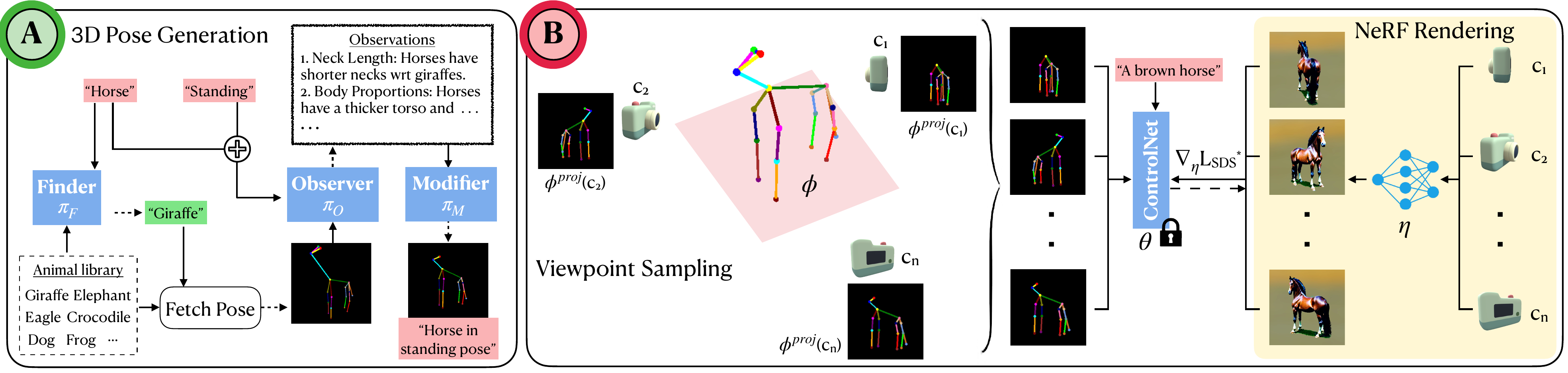}
    \caption{\textbf{Automatic pipeline for 3D animal generation.} Given the name of an animal and textual pose description, we utilize a multi-agent LLM to generate a 3D pose ($\phi$) supported by a small library of animal names paired with 3D poses. With the obtained 3D pose, we train a NeRF to generate the 3D animal guided by a diffusion model controlled by 2D views ($\phi^{proj}$) of $\phi$.}
    \label{fig:llm_poseedit}
    \vspace{-3mm}
\end{figure}

\section{Method}

Multi-view sampling from T2I diffusion models for 3D generation is guided using directional prompt extensions such as ``<user\_text>, front view'' and ``<user\_text>, side view''. Such a control signal is ambiguous due to, 1) directional text remaining unchanged over a range of camera parameters, 2) T2I diffusion models generating deformable entities in various poses for the same view. Thus we utilize 3D pose as a stronger guidance to maintain consistency over different views. To do this in a 3D consistent manner we design 1) a model to generate 2D image samples following the projection $\phi^{proj}$ of a 3D pose $\phi$ of an animal, 2) a method to generate the 3D pose $\phi$ of a novel animal $y$ using a limited library of 3D poses (${\Phi}$) of commonly observed animals in nature and a multi-agent LLM pose editor, and 3) a method to create 3D animals given an animal name $y$ and 3D pose $\phi$. Our 3D model is represented using Neural Radiance Fields (NeRF~\cite{mildenhall2021nerf}).

\subsection{TetraPose ControlNet}
To train a model to follow pose control we require images of animals with annotated poses. Datasets released by ~\cite{banik2021novel} and ~\cite{ng2022animal} provide 2D pose annotations of animal images spanning a large number of species, compared to the limited diversity available in 3D animal pose datasets~\cite{xu2023animal3d, badger20203d}. We thus utilize these 2D pose datasets by learning to map the 2D pose of an animal to its captured image. We define such datasets of animal species $y_j$, corresponding animal images $x_j$, and their 2D pose $\phi^{proj}_j$ as the set $\mathcal{D}=\{(x_j,\phi^{proj}_j, y_j)\}_{j=1}^J$, where $J = |\mathcal{D}|$ is the number of image-pose pairs in the dataset. This learned mapping can then be used to generate multi-view image samples consistent with a 3D pose $\phi$. The mapping is represented by a ControlNet that produces animal images across mammals, amphibians, reptiles and birds following a 2D input pose condition $\phi^{proj}_j$ learned by minimizing the following objective:
\begin{equation}
     \mathcal{L}_{ControlNet} = \mathbb{E}_{z_0, t, y_j, \phi^{proj}_j, \epsilon \sim \mathcal{N}(0,I)}\left[ \|\epsilon - \epsilon_\theta(z_t; t, y_j, \phi^{proj}_j)\|^2 \right],
\end{equation}
where $z_0 = x_j$. The above objective aims to learn a network $\epsilon_\theta$ to estimate the noise added to an input image $z_0$ (or $x_j$) to form a noisy image $z_t$ given time-steps $t$, text $y_j$ and pose condition $\phi^{proj}_j$. The network $\epsilon_\theta$ is represented by the standard U-Net architecture of diffusion models (Stable Diffusion in this case) with a trainable copy of the U-Net's encoder attached to it using trainable zero convolution layers. We provide training details in Sec.~\ref{sec:implementation_details_suppl}.

% The trained TetraPose-ControlNet can be used to generate pose controlled images of tetrapods -- mammals, reptiles, birds, and amphibians. The model performs well for out-of-domain 2D pose inputs of animals not seen during training. The model also performs well for inputs consisting of modified 2D poses with extra appendages such as multiple heads, limbs, wings, and/or tails. 

The trained TetraPose ControlNet can be used to generate pose-controlled images of tetrapods, including mammals, reptiles, birds, and amphibians. The model performs well for out-of-domain 2D pose inputs of animals not seen during training. It also performs well with inputs consisting of modified 2D poses that include extra appendages such as multiple heads, limbs, wings, and/or tails. While T2I diffusion models inherently provide a huge diversity to the generated outputs, the control module provides strong controlling signals to generate appropriate body parts in the right positions, alleviating the problem of T2I diffusion models producing inconsistent multi-view images when prompted using directional texts only.

% As the training consists of natural images captured in-the-wild, they contain heavy self-occlusions leading to absence of keypoints and/or absence of bones between keypoints in the 2D pose images $\phi^{proj}_j$. Consequently training the model using such samples help associate each keypoint and bone with its corresponding body part in the image, rather than the complete instance of the animal.
\begin{figure}
    \centering
    \includegraphics[width=0.9\linewidth]{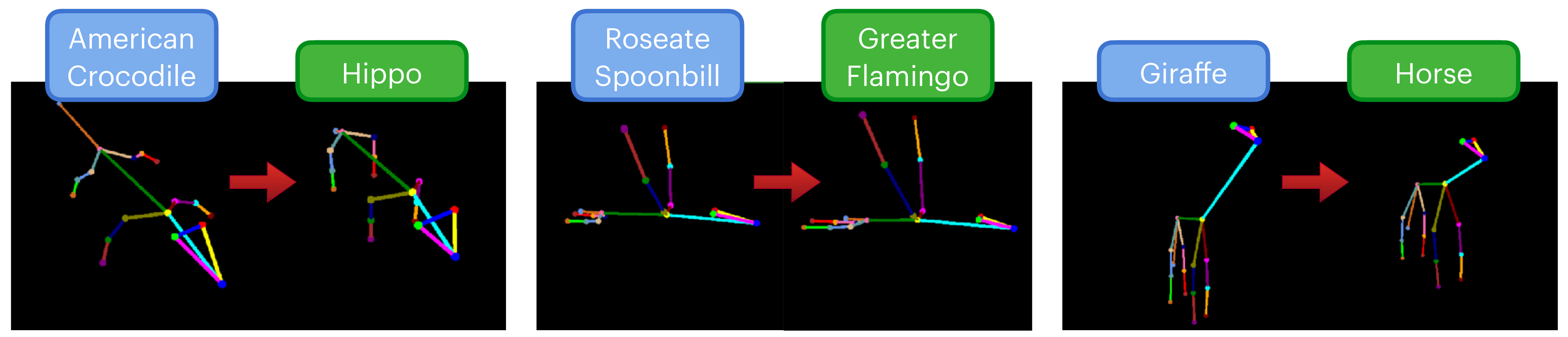}
    \caption{\textbf{Qualitative examples of pose editing using multi-agent LLM setup.}  For each example, the green box denotes the desired animal, while the blue box is the animal retrieved from the 3D pose library by Finder LLM ($\pi_{F}$). We show the pose modification performed by the joint effort of Observer ($\pi_{O}$) and Modifier ($\pi_{M}$) for three instances.}
    \label{fig:llm_poseedit}
    % \vspace{-3mm}
\end{figure}

\subsection{3D Pose Generation aided by Multi-agent LLM}
Generating a 3D pose based on a text is not trivial as text-to-pose is a many-to-many mapping. Existing 3D animal pose datasets are not diverse or vast enough to learn this mapping for a variety of animals. Thus we leverage LLMs which are pre-trained on expansive textual datasets, and thus can reason about anatomical proportions of various animals. We find that LLMs do not produce good 3D poses using only a text input, instead we use LLMs to adapt a input 3D pose to represent a novel animal. We created a limited library consisting of 16 animal 3D poses for this purpose.

Given a library of animals $\mathcal{B}=\{(y_i,\phi_i)\}_{i=1}^n$ consisting of 3D keypoint positions $\phi_i \in {\Phi}$ and animal names $y_i \in {\cal Y}$, we utilize a multi-agent LLM setup for creating a 3D pose for any desired animal $y$ and pose description $p$. The agents include 1) Finder ($\pi_{F}$), 2) Observer ($\pi_{O}$), and 3) Modifier ($\pi_{M}$). Let the keypoint names representing any animal be $\mathcal{K}$ and let the bone sequence which defines the skeleton be $\mathcal{S}$. Given $\mathcal{K}$, the Finder selects the animal in $\mathcal{B}$ that is ``anatomically closest'' to the desired animal $y$ as $(y_c,\phi_c) = \pi_{F}(y, \mathcal{B}, \mathcal{K})$. ``Anatomically closest'' is defined as the animal whose 3D pose will require minimal modifications/updates to represent $y$. Given the keypoint definitions $\mathcal{K}$, bone sequence $\mathcal{S}$, the desired animal name $y$, the animal $y_c$ selected by $\pi_{F}$, and the pose description $p$, the Observer generates $\mathcal{O} = \pi_{O}(y_c, y, p, \mathcal{S}, \mathcal{K})$. $\mathcal{O}$ represents a plan describing which keypoints of $y_c$ should be adjusted along with a set of instructions for the Modifier to implement the suggested adjustments to represent the 3D pose of the desired animal $y$ in the described pose $p$. Based on the observations $\mathcal{O}$, the Modifier updates the 3D positions of the keypoints, $\phi_c$, of the closest animal to $\phi = \pi_{M}(\phi_c, \mathcal{O})$. Thus we obtain the 3D keypoint positions $\phi$ of the desired animal $y$ in the described pose $p$. We find that this multi-agent procedure is more stable and accurate than using a single LLM for pose generation (see Sec.~\ref{sec:Additional_Ablations_suppl}). We are able to represent diverse animals observed in nature using this setup. Fig.~\ref{fig:llm_poseedit} presents examples of pose editing using our described setup. As ground truth text-to-3D poses for animals do not exist and the described task is a many-to-many problem, quantitative evaluation is difficult to obtain. Thus we conducted a user study to evaluate the efficacy of our method (details in Sec.~\ref{sec:experiments}). We describe the contents of our library $\mathcal{B}$ and the prompts to LLMs in detail in Sec.~\ref{sec:animal_library_suppl} and Sec.~\ref{sec:multi-agent_LLM_implementation_details_suppl}. 

\subsection{Pose Editor and Shape Initializer} To facilitate easy creation and editing of 3D poses, we present a user-friendly tool to modify, add, or delete joints and bones. We also provide a method in this tool to automatically generate an initial shape based on the 3D skeleton using simple 3D geometries such as cylinders, cones, and ellipses. We use this shape to pre-train our NeRF before fine-tuning using diffusion based guidance. Details of this tool are presented in Sec.~\ref{sec:implementation_details_suppl}.

\subsection{Bringing Bones to Life}
We want to create 3D animals given an input text $y$ and 3D pose $\phi$. We adopt the Score Distillation Sampling (SDS) method proposed in DreamFusion~\cite{poole2022dreamfusion}, adapted for our TetraPose ControlNet. The SDS loss gradient can now be represented as:
 \begin{equation}
  \nabla_\eta \mathcal{L}_{\text{SDS}}(\theta, z = \mathscr{E}(g(\eta, c))) = \mathbb{E}_{t, c, \epsilon}\left[ w(t) \left(\epsilon_\theta(z_t; t, y(c), \phi^{proj}(c)) - \epsilon\right) \frac{\partial z}{\partial \eta} \right], 
\end{equation}
 where $\eta$ represents the trainable parameters of the NeRF, $\theta$ the frozen diffusion model parameters, $c$ represents the sampled camera parameters, $t$ is the number of time-steps, and $w(t)$ is a timestep-dependent weighting function. $z$ denotes the latent encoded using encoder $\mathscr{E}$ for the image rendered from the NeRF $g(\eta, c)$ for camera $c$. $y(c)$ represents the directional text created based on camera $c$, while $\phi^{proj}(c)$ is the 2D projection of the 3D pose $\phi$ for camera $c$. Additionally, we also utilize an image domain loss weighted by the hyper-parameter $\lambda_{RGB}$ which reduces flickering and produces more solid geometry (Fig.\ref{fig:lrgb}):
 \begin{equation}
 \mathcal{L}_{RGB} = \lambda_{RGB} \cdot \mathbb{E}_{t, c, \epsilon}\left[ w(t) \cdot \| g(\eta, c) - \mathscr{D}(\hat{z}) \|^2\right],
\end{equation}
where $g(\eta, c)$ is the image rendered from the NeRF and $\mathscr{D}(\hat{z})$ is the denoised image decoded using decoder $\mathscr{D}$ from the denoised latent $\hat{z}$.

Since our TetraPose ControlNet is trained on much smaller number of images compared to Stable Diffusion, it loses diversity. To improve diversity and generation capability, we propose to use control scheduling and guidance scheduling. We observe that higher control scale provides strong signal for geometry modeling whereas higher guidance scale provides strong signal for appearance modeling. Since geometry is perfected in the initial stages and appearance in the latter, we propose reduction of control scale and increase of guidance scale over training iterations. This helps us create out-of-domain assets with significant style variety (see Fig.~\ref{fig:style_prompts}). Our strategy is formulated as:
\begin{equation}
 \text{control\_scale}  = \cos(\frac{\pi}{2} \cdot \frac{\text{train\_step}}{\text{max\_step}}) \cdot (\text{control}_{max} - \text{control}_{min}) + \text{control}_{min}, 
\end{equation}
 
\begin{equation}
 \text{guidance\_scale}  = \frac{\text{train\_step}}{\text{max\_step}} \cdot (\text{guidance}_{max} - \text{guidance}_{min}) + \text{guidance}_{min}, 
\end{equation}

where $\text{train\_step}$ is the current training step and $\text{max\_step}$ is total training iterations. The variables $\text{control}_{max}$, $\text{control}_{min}$, $\text{guidance}_{max}$, $\text{guidance}_{min}$ are hyperparameters. We show that a linear scheme is better for guidance scheduling, while cosine is better for control scheduling in Sec.~\ref{sec:Additional_Ablations_suppl}.

\begin{figure}
    \centering
    \includegraphics[width=1\linewidth]{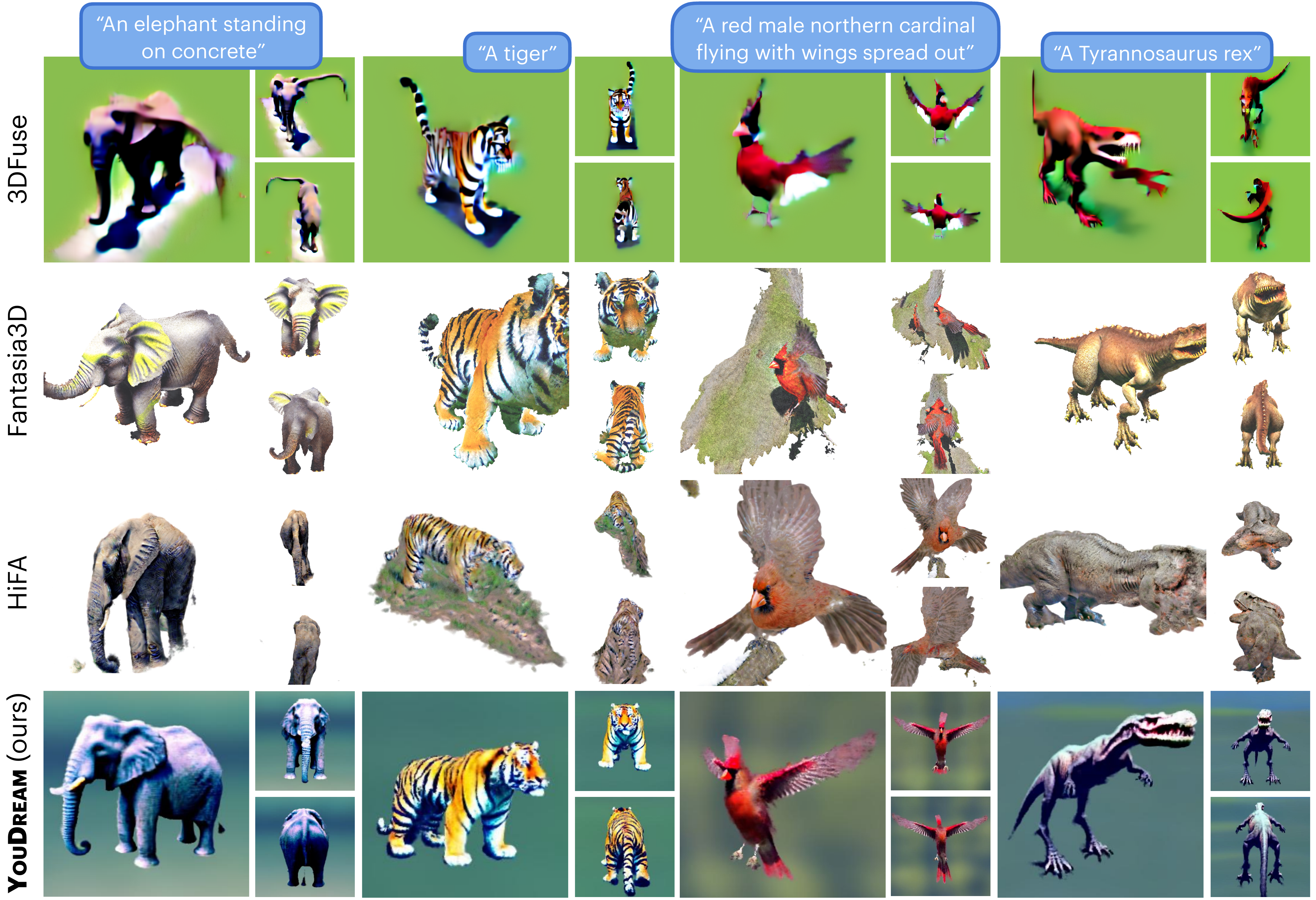}
    \caption{\textbf{Comparison on generating animals observed in nature.} We compare with baselines which use T2I diffusion (with official open-source code) for the automatic generation of text-to-3D animals. Unlike the baselines, our method produces high quality anatomically consistent animals.}
    \label{fig:common_animal}
    % \vspace{-2mm}
\end{figure}

\section{Experiments}
\label{sec:experiments}
In this section, we compare \textsc{YouDream} against various baselines and evaluate the effect of various components of our method. We show qualitative comparison with text-to-3D methods which are guided by T2I diffusion models for common animals observed in nature. We also compare with MVDream -- which uses a (text + camera)-to-image diffusion model trained on 3D objects. It should be noted that our method does not use any 3D objects for training, yet is able to deliver geometrically consistent results. We conduct a user study to quantitatively evaluate our method against these baselines. We also compute CLIP score following previous work~\cite{shi2023mvdream}, shown in Sec.~\ref{sec:clip_evaluation_suppl}. Additionally, we present ablations over the various modules that constitute \textsc{YouDream}. 

\textbf{Generating animals observed in nature.} In Fig.~\ref{fig:common_animal}, we compare our method against 3DFuse~\cite{seo2023let}, Fantasia3D~\cite{chen2023fantasia3d}, and HiFA~\cite{zhu2023hifa} for generating common animals. HiFA and Fantasia suffer from anatomical inconsistency, while 3DFuse is more consistent in some cases due to the use of depth control. However, 3DFuse is highly dependent on the point cloud prediction leading to the generation of implausible geometry, for example in the case of elephant and T-Rex. It should be noted that generating results using Fantasia3D required extensive parameter tuning, which has also been indicated by the authors in their repository. All results are generated using the same seed 0 for fair comparison. We use the default hyperparameter settings of each baseline except Fantasia3D. The text ``, full body'' is appended at the end of the prompt for all baselines, as we observed that the methods generate truncated animals in many cases. We generate common animals using our fully automated pipeline, where we use LLM for pose editing sourced by a library of 3D poses. Tiger is generated by our multi-agent LLM based on a German Shepherd, northern cardinal is made from an eagle, while elephant and Tyrannosaurus rex are part of our library. In all cases, our method visibly outperforms baselines in terms of perceptual quality and 3D consistency.

\begin{wrapfigure}{t}{0.45\columnwidth}
\vspace{-5mm}
    \centering
    \begin{subfigure}{0.4\columnwidth}
        \centering
        \includegraphics[width=\linewidth]{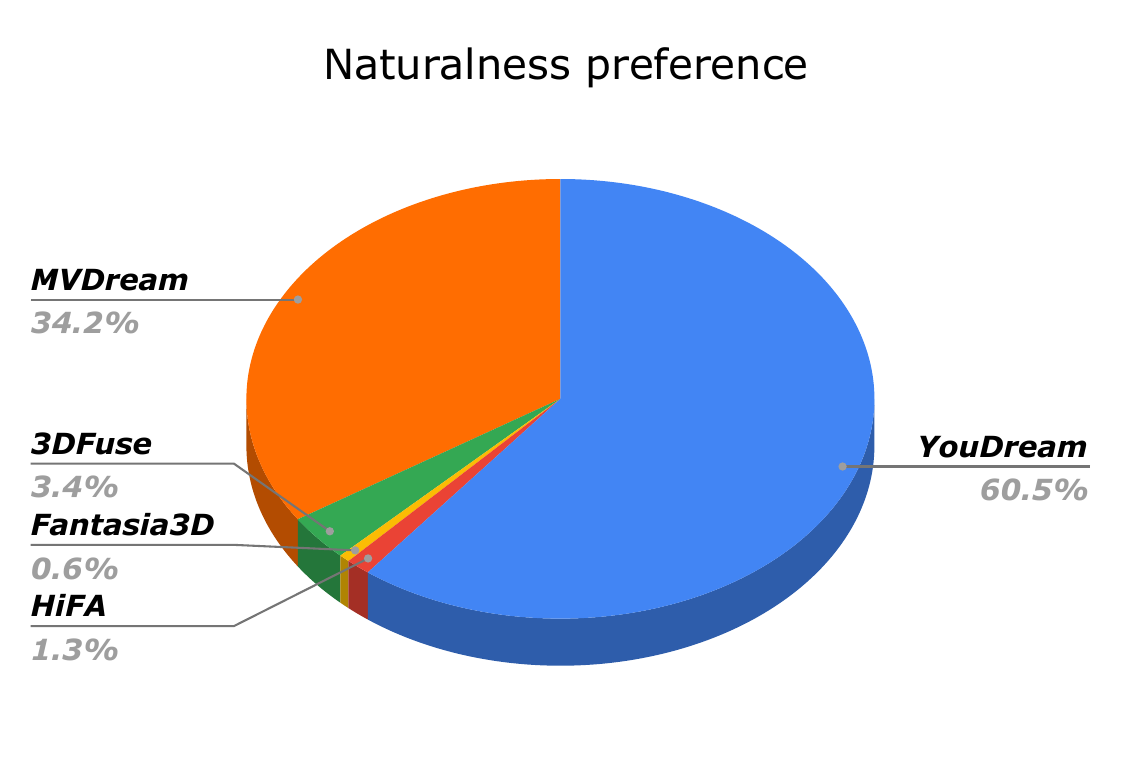}
        \label{fig:user_study_naturalness}
    \end{subfigure}
    \vfill
    \vspace{-5mm}
    \begin{subfigure}{0.4\columnwidth}
        \centering
        \includegraphics[width=\linewidth]{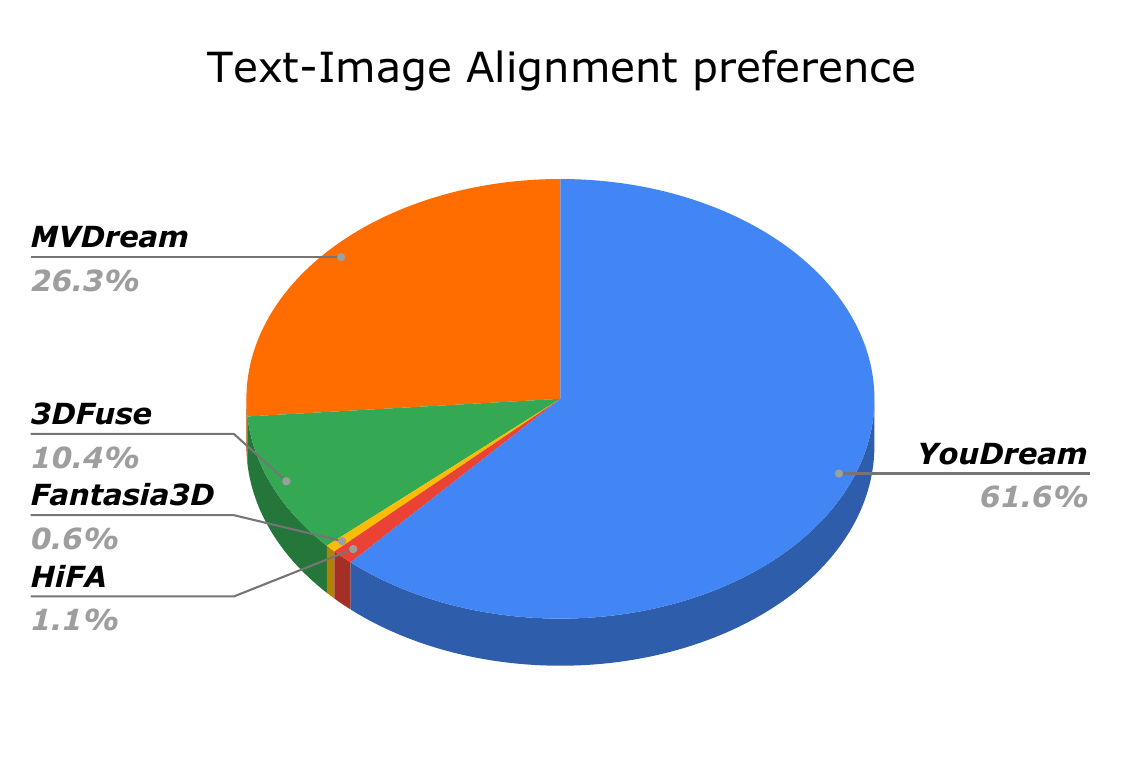}
        \label{fig:user_study_T2I}
    \end{subfigure}
    \vspace{-5mm}
    \caption{\textbf{User Study.} User preferences on 1) Naturalness and 2) Text-Image alignment averaged over 32 participants and 22 text-to-3D generated assets reveals the superiority of our proposed method.}
    
    \label{fig:user_study}
\end{wrapfigure}

\textbf{Generating unreal creatures.} A major advantage of our pipeline is that it can be easily used to generate non-existent creatures, especially those not explainable through text. These can be generated robustly using our method when the user provides a skeleton of their concept. We use our pose editor tool to generate the results shown in Fig.~\ref{fig:creative_creation}, where \textsc{YouDream} produces stunning unreal creatures. We show the pose controls we use in Sec.~\ref{sec:implementation_details_suppl}. Notably, MVDream~\cite{shi2023mvdream} struggled to follow the textual prompt as such creatures are not represented in existing 3D datasets; producing incorrect results such as ``Wampus cat'', a cat-like creature in American folklore, with three legs instead of six in Fig.~\ref{fig:creative_creation} row 5. In some aspects HiFA attempted to follow the prompt (owing to its usage of T2I Stable Diffusion) such as producing a couple of tentacles in Fig.~\ref{fig:creative_creation} row 1 and more than one head in row 3, but produces geometrically inconsistent results in all cases. Again we use seed 0, default hyperparameter settings for baselines and append ``, full body'' at the end of the prompts except for `golden ball'.

    % 1. include ablations:
    %     a) Latent-nerf vs latent-tune
    %     b) latent size
    %     c) start with init, no control; without init, with control; without init, without control
    % 2. compare with:
    %     3Dfuse, HiFA, Fantasia3D, MVDream
    % 3. Show a) Crazy Control Pose; b) Control makes better Asset
    % 4. compositing, unnatural body shapes
    % 4. Out-of-Domain: mermaid maybe?
    % 5. Give codes for HIFA, DreamFusion so that reviewers can see what HiFA generates.
    % 6. show evidence of various LLMs for pose editing
    % 7. compare with 3DFauna / LASSIE
    % 8. start with SMAL, start with balloon animal
    % 9. without and with various annealings
    % 10. change shape of human skeleton using LLM
    % 11. complex prompts such as wearing stuff, standing on stuff

\textbf{Subjective Quality Analysis.} We conducted a voluntary user study with 32 participants to subjectively evaluate the quality of our 3D generated assets. The participants were shown side-by-side videos of assets generated using the same prompt input by \textsc{YouDream}(ours), HiFA, Fantasia3D, 3DFuse, and MVDream, and were asked to select the best model under the categories - 1) Naturalness and 2) Text-Image alignment. The participants were instructed to judge naturalness on the basis of geometrical and anatomical consistency/correctness, perceptual quality, artifacts, and details present in the videos. Text-image alignment preference was self-explanatory. A total of 22 prompts and their corresponding 3D assets generated using each model were shown to each participant, thereby accumulating a total of 1408 user preferences. Of the 22 prompts, 13 involved naturally existing animals while the remaining 9 included unreal and non-existent animals. The collected user preferences are shown in Fig.\ref{fig:user_study}. We observe a 60-62\% user preference in both the preference categories for our model, strongly indicating the superior robustness and quality of \textsc{YouDream}. 

We also tested the efficacy of our multi-agent LLM based pose generator via a subjective study. We request 16 novel 3D poses of different animals from the multi-agent LLM which uses 16 pre-defined animal poses in our animal pose library. The requested animals were such that there was a high chance of using each reference animal pose in the library. The participants were shown paired videos of rotating 3D poses, consisting of the pose taken from the library (left-side video, `reference animal') and the novel pose (right-side video, `requested animal') generated. Since the participants were not experts in animal anatomy, they were also provided multi-view images of each animal under their video. They were asked to mark `Yes' or `No' for the question: ``If this 3D pose <reference pose video> represents `reference animal' in `reference pose' pose. Could this 3D pose <generated pose video> represent `requested animal' in `requested pose' pose?'' The study consisted of the same 32 participants and each subject voted for all the 16 novel poses. Subjects agreed that the generated pose correctly represents `requested animal' 91\% of the time. Kangaroo (standing pose) generated by the multi-agent LLM using the pose of T-Rex (standing pose) received the lowest agreement among all pairs with 8 out of 32 votes being `No'. Detailed description of the pose library, the generated poses, and particulars of the human study are provided in Sec.~\ref{sec:user_study_suppl}.  

\begin{figure}
    \centering
    \includegraphics[width=0.95\linewidth]{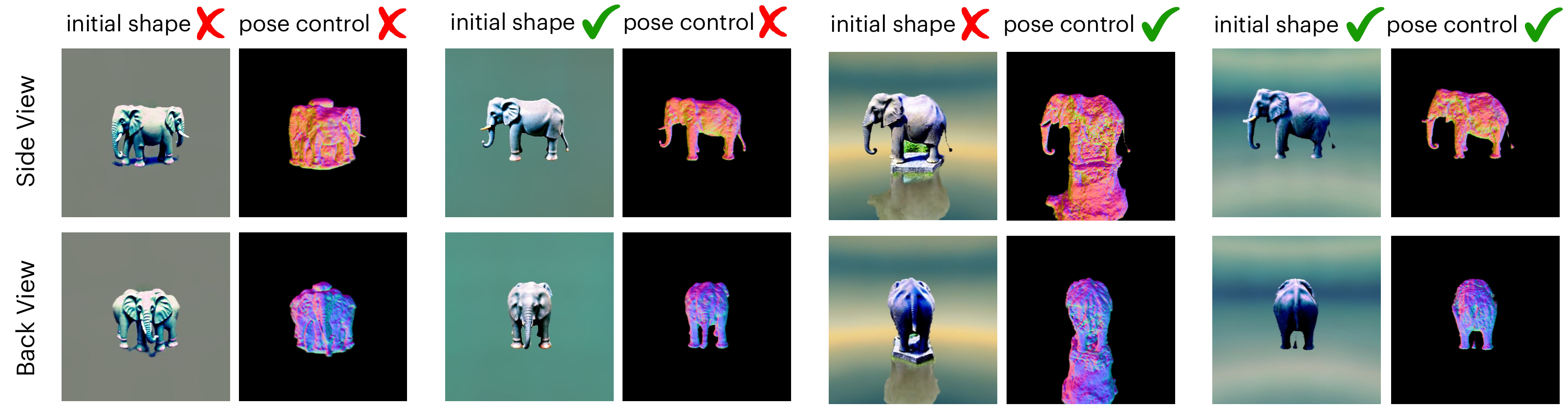}
    \caption{\textbf{Ablation over the effect of initial shape and pose control.} The initial shape helps in producing clean geometry, while the pose control helps to maintain 3D consistency. }
    \label{fig:ablation_init_and_pose_control}
\end{figure}

\textbf{Ablation.} We present ablation over the effect of using initial shape and pose control in Fig.~\ref{fig:ablation_init_and_pose_control}. Without pose control refers to using vanilla Stable Diffusion. Without using initial shape or control, the Janus head problem occurs. With initial shape but without pose control, the geometry improves but still sees the appearance of another head on the elephant's backside. Using pose control without initial shape produces visibly good results, however using both initial shape and pose control results in much cleaner geometry.

\begin{figure}
    \centering
    \includegraphics[width=0.95\linewidth]{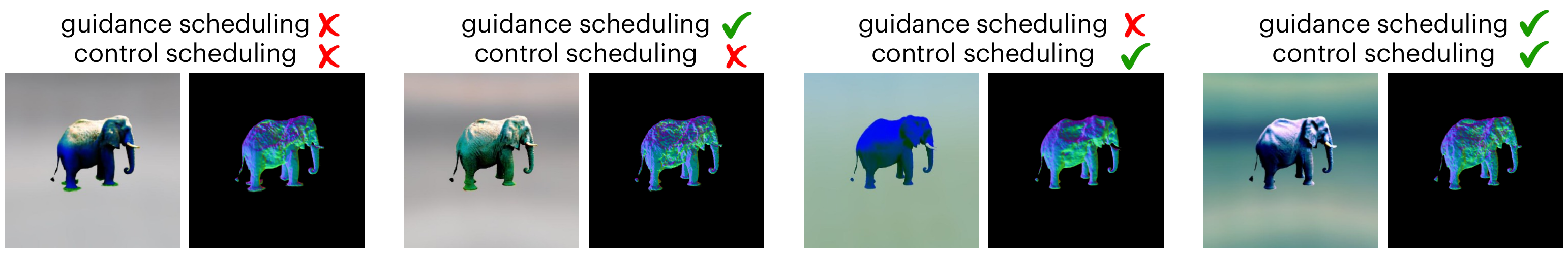}
    \caption{\textbf{Ablation over scheduling techniques.} Using either guidance or control scaling produces unnatural color, using neither produces artifacts such as grass at feet owing to lower diversity of ControlNet compared to Stable Diffusion.}
    \label{fig:ablation_scheduling_control}
    % \vspace{-3mm}
\end{figure}

In Fig.~\ref{fig:ablation_scheduling_control} we show the effect of our scheduling strategies. Without guidance or control scaling, the result has grassy texture at the feet which could be owing to seeing most elephants on grass during TetraPose ControlNet training on limited animal pose data. Using only one kind of scheduling produces incorrect color, showing that both scaling techniques go hand-in-hand. 

\section{Conclusion}
We presented \textsc{YouDream}, a method to create anatomically controllable and geometrically consistent 3D animals from a text prompt and a 3D pose input. Our method facilitates the generation of diverse creative assets through skeleton control, which cannot be expressed through language and is difficult to provide as guidance image, especially for unseen creatures. Additionally, we presented a pipeline for automatic generation of 3D pose for animals commonly observed in nature by utilizing a multi-agent LLM setup. Our 3D generation process enjoys multi-view consistency by utilizing a 3D pose as a prior. We quantitatively outperform prior work in terms of ``Naturalness'' and ``Text-Image Alignment'' as evidenced in the user study.

% LET 2D DIFFUSION MODEL KNOW 3D-CONSISTENCY FOR ROBUST TEXT-TO-3D GENERATION
% - text-to-image is inconsistent many times, that is used for creating control resulting in anatomical inconsistencies

% Section in related: 3D consistency in text to 3D using SDS:
% 3DFuse
% DreamPolisher
% MVDream
% SweetDreamer
% DreamWaltz, AvatarCraft, DreamAvatar

% three headed dragon, narasimha, kaliya, chimera, 

% varioius seeds

% adding wings on ball

% discuss about camera pose or animal pose vague using directional prompts

% show following pose suppl

% supplemetary show mvdream for common animals

% put zor on fact that our training does not use any 3d models but yet perform better than mvdream

{
    \small
    \bibliographystyle{ieeenat_fullname}
    \bibliography{main}
}

%%%%%%%%%%%%%%%%%%%%%%%%%%%%%%%%%%%%%%%%%%%%%%%%%%%%%%%%%%%%
\newpage
\appendix
\noindent{\Large \textbf{Appendix}}

\section{More Results}
\label{sec:more_results_suppl}
Check our webpage for video results at this \textcolor{blue}{\href{https://youdream3d.github.io/}{URL}}.

\begin{figure}[h]
    \centering
    \includegraphics[width=1\linewidth]{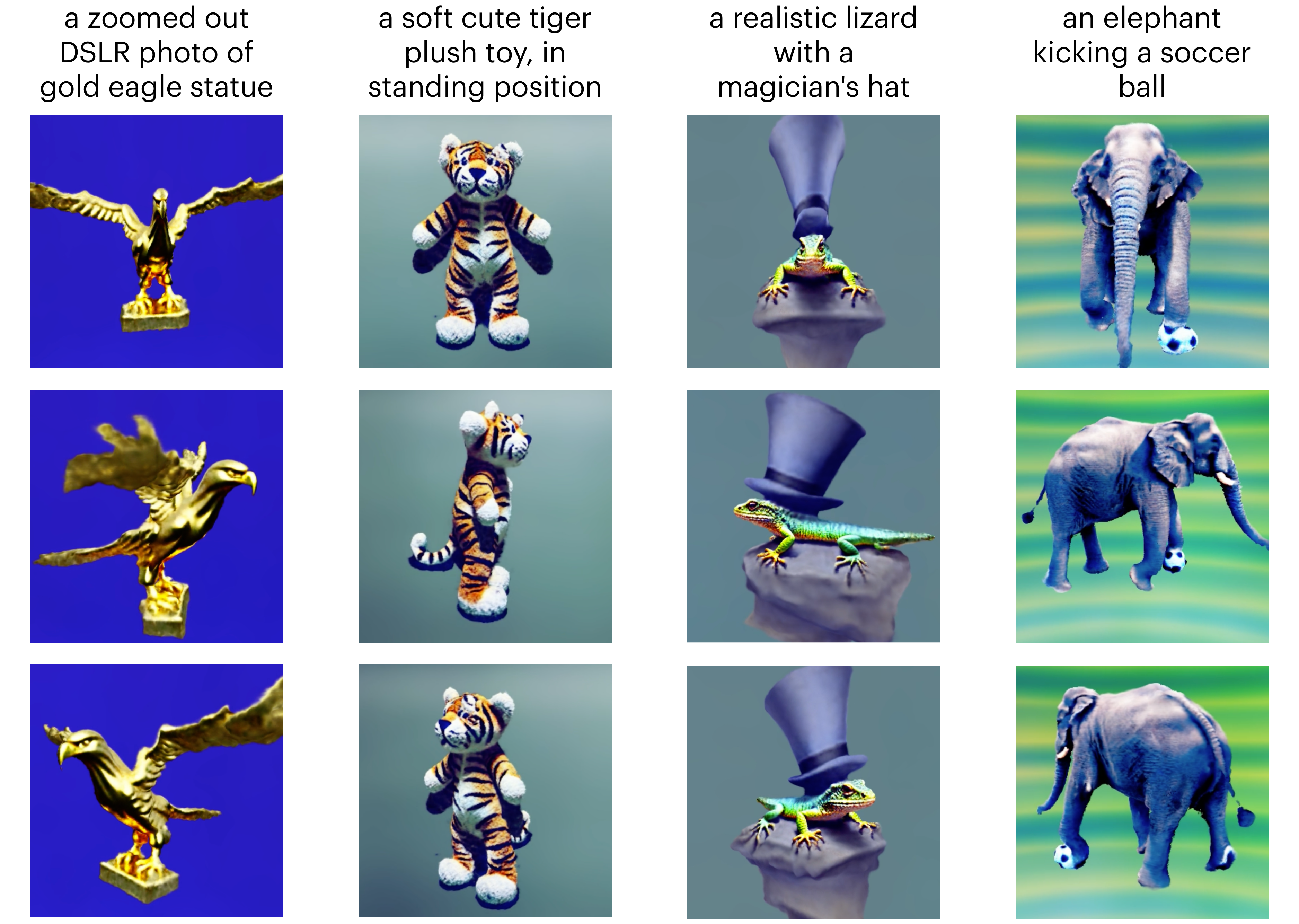}
    \caption{\textbf{Results on compositional and style prompts.} We show our method performs well while generating animals with style alterations or object interactions.}
    \label{fig:style_prompts}
\end{figure}

We show our method's performance for varied styles in Fig.~\ref{fig:style_prompts}. Even though our ControlNet is trained on images of animals in the wild, \textsc{YouDream} produces assets with significant style alteration. This is attainable because of our control scheduling and guidance scheduling approach, which ensures consistent geometry to be formed in initial iterations with higher control scale and style being finalized during the later iteration with higher guidance scale.

\begin{figure}[h]
    \centering
    \includegraphics[width=1\linewidth]{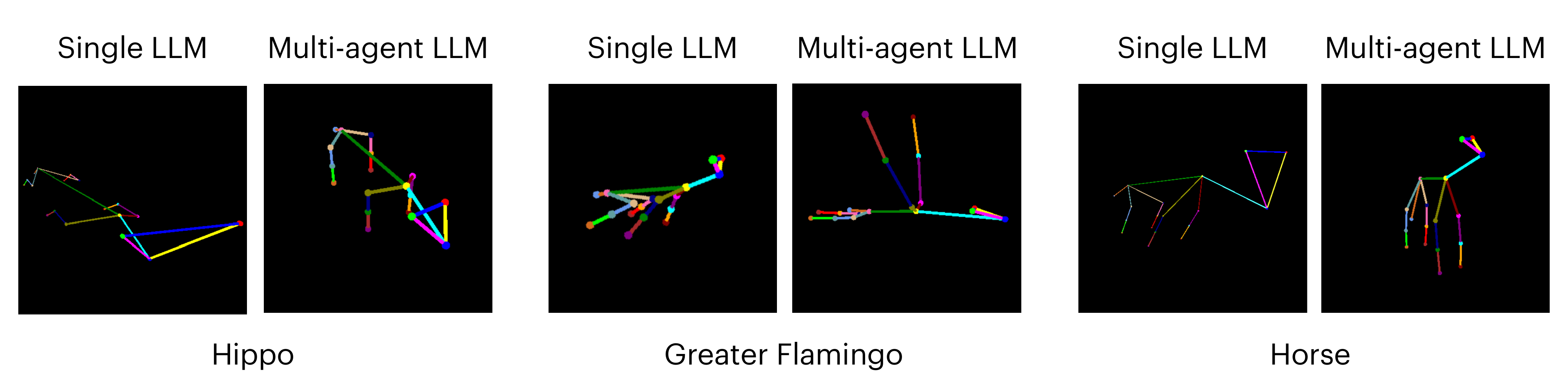}
    \caption{\textbf{Pose generation using single LLM vs our multi-agent LLM setup.} For ``Hippo'', ``Greater Flamingo'', and ``Horse'', we show a 2D view of the 3D pose generated by a single LLM compared to our multi-agent setup.}
    \label{fig:singlevsmulti}
\end{figure}

\begin{figure}[h]
    \centering
    \includegraphics[width=1\linewidth]{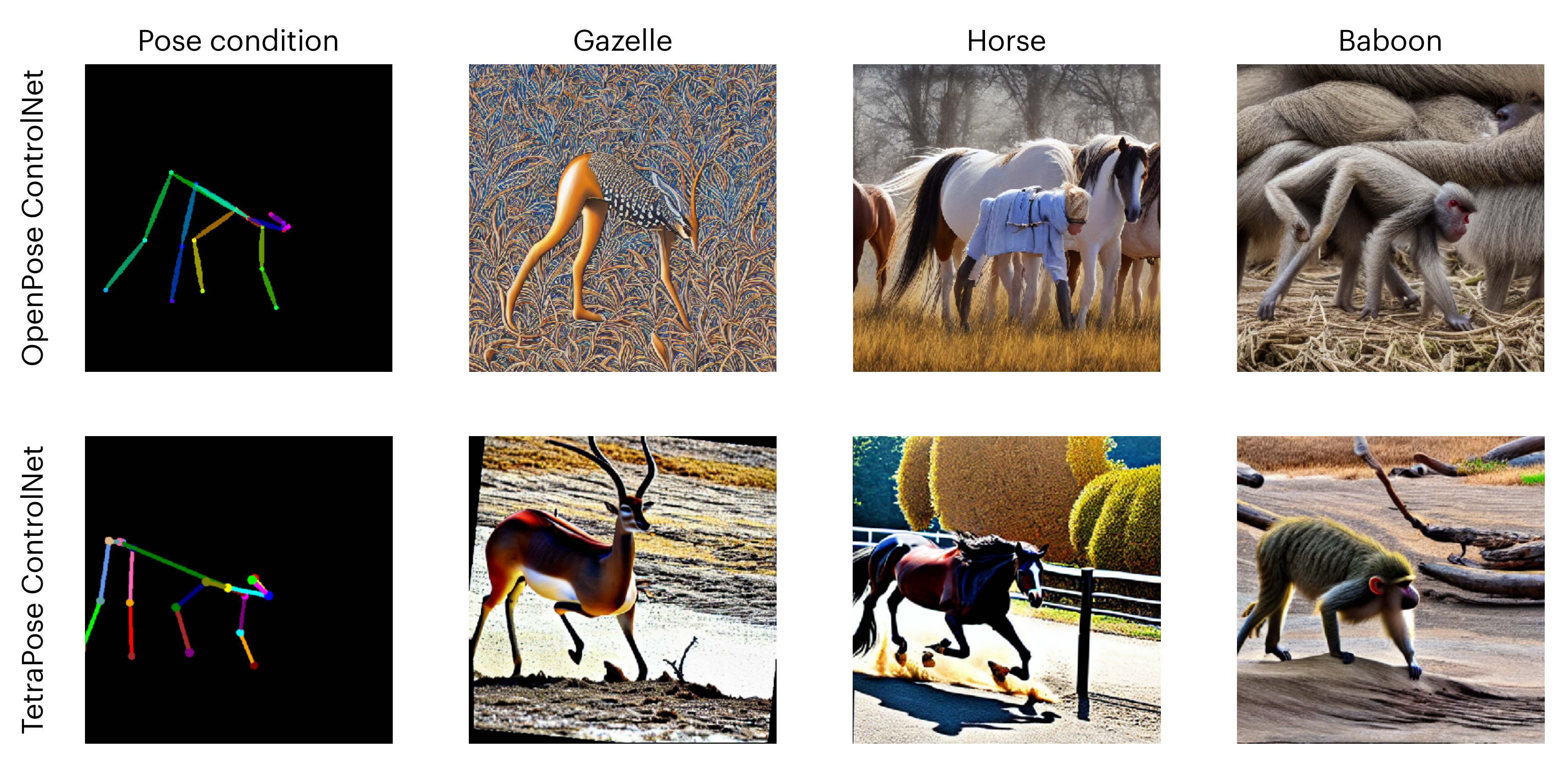}
    \caption{\textbf{Comparison with OpenPose ControlNet for generating animals.} OpenPose ControlNet produces the animal in the prompt for ``Horse'' and ``Baboon'', but either does not follow control or makes unnatural anatomy. For ``Gazelle'' a meaningless image is produced.}
    \label{fig:compare_ControlNet}
\end{figure}

\begin{figure}[h]
    \centering
    \includegraphics[width=1\linewidth]{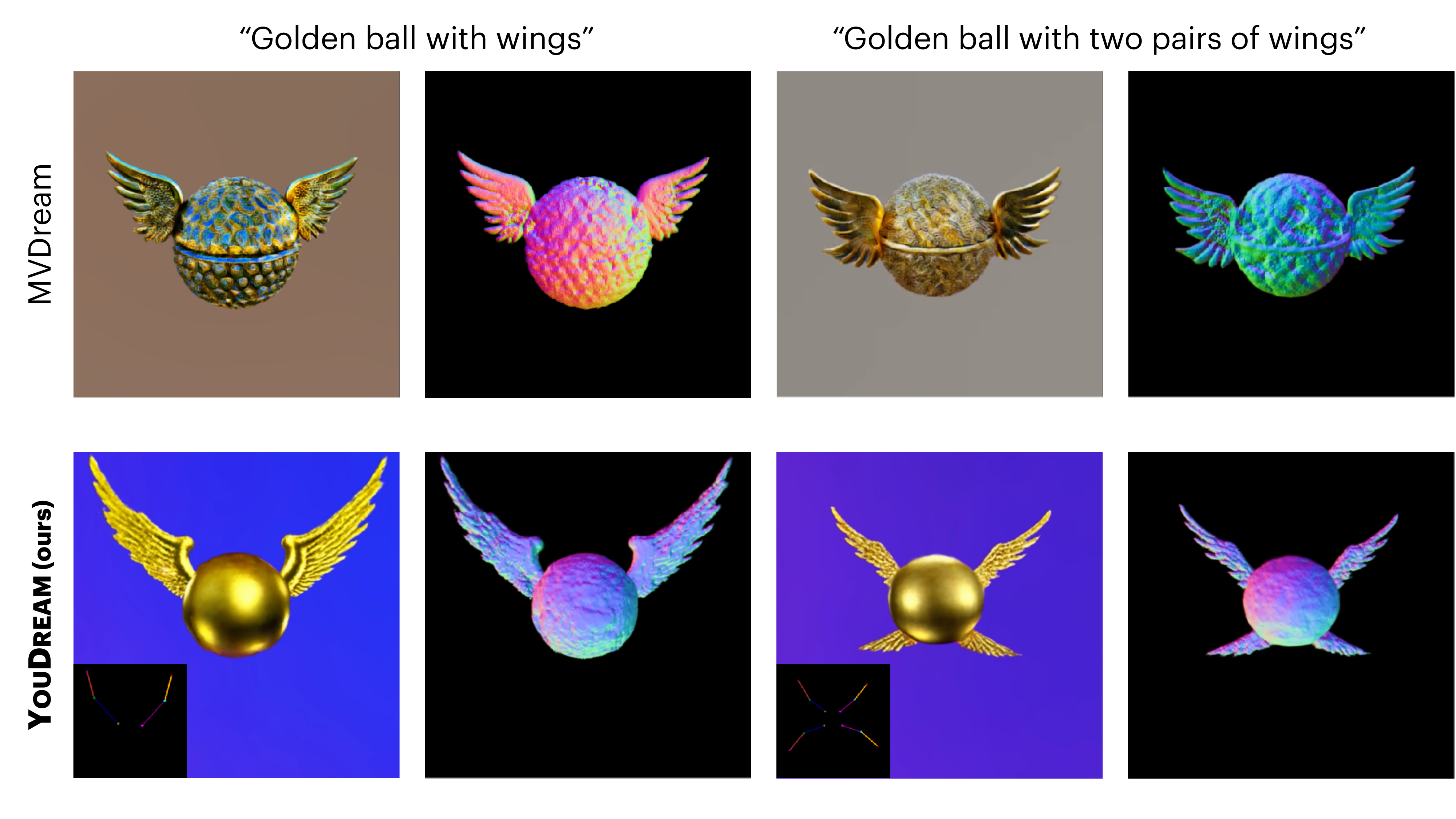}
    \caption{\textbf{Toy example showing inefficacy of text prompt.} We show that pose control helps to add the additional wings at the desired location. MVDream makes two wings for ``Golden ball with two pairs of wings'' with differently shaped wings compared to ``Golden ball with wings''.}
    \label{fig:ball_compare}
\end{figure}

\begin{figure}[h]
    \centering
    \includegraphics[width=1\linewidth]{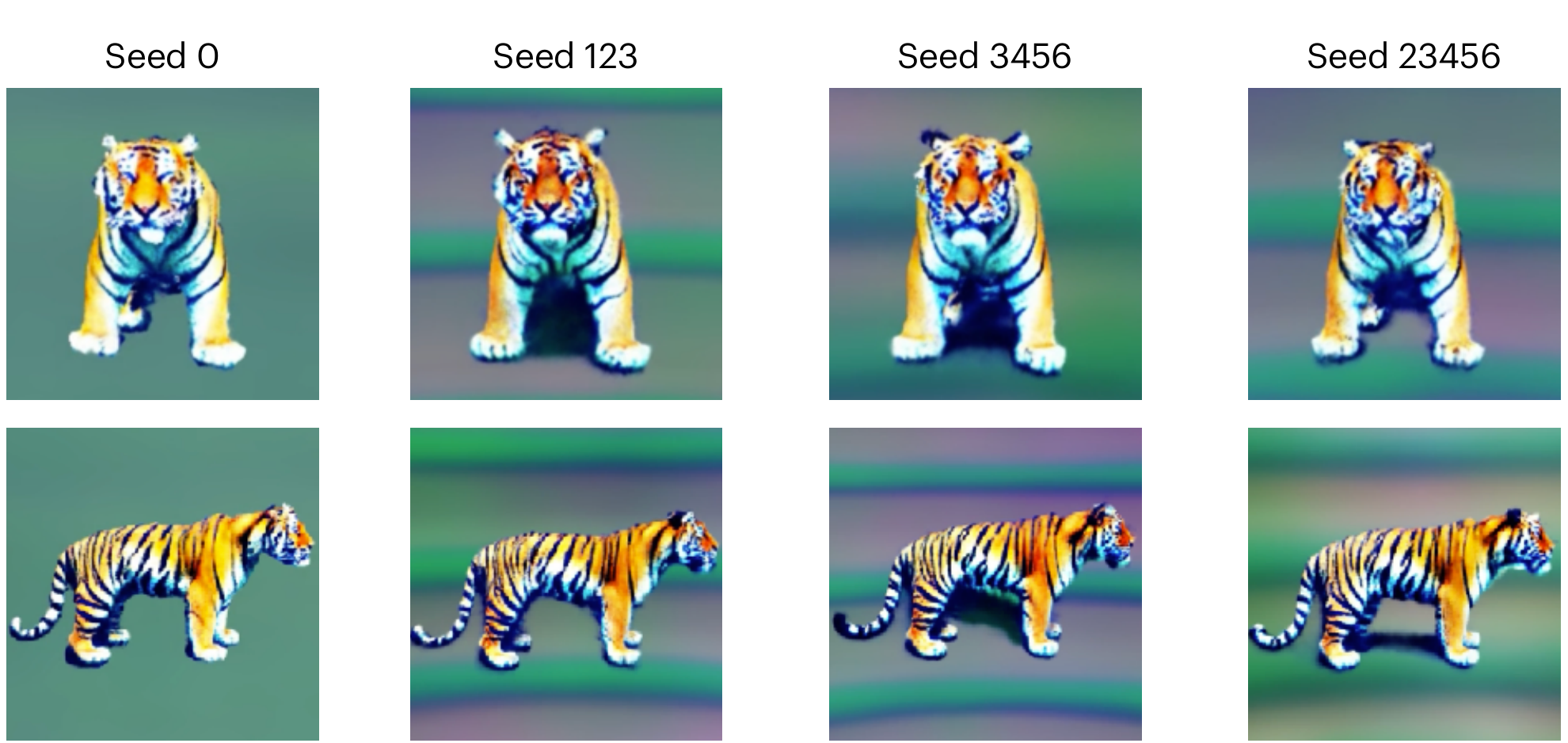}
    \caption{\textbf{Variation with seed.} Our method is robust across seeds and generates slightly different faces and stripes for various seeds.}
    \label{fig:seed_vary}
\end{figure}

\begin{figure}[h]
    \centering
    \includegraphics[width=1\linewidth]{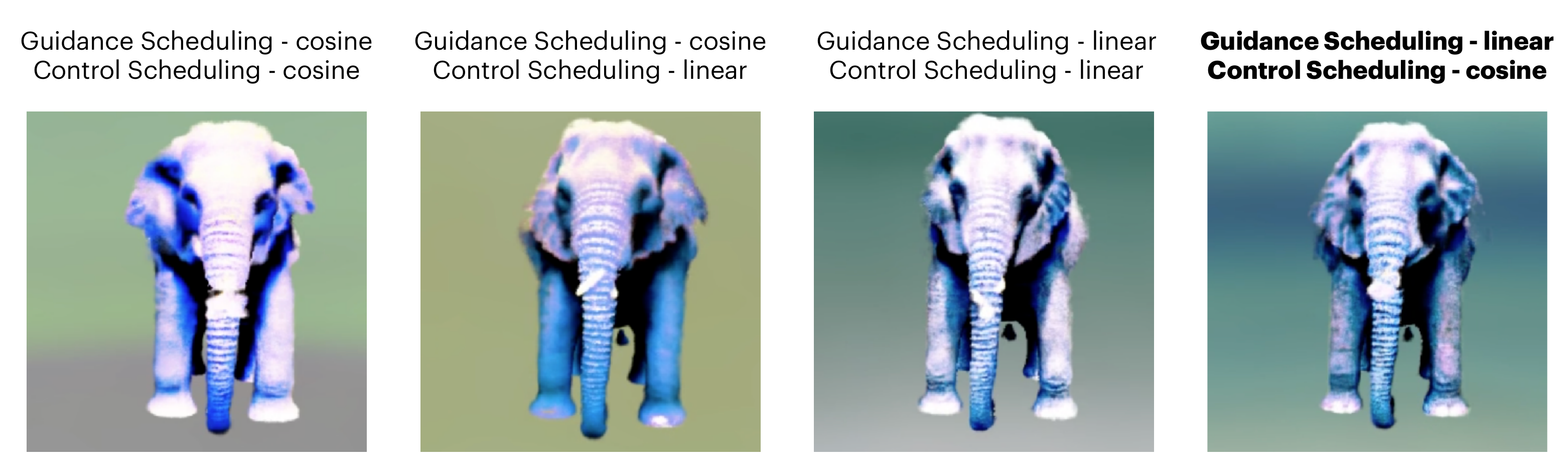}
    \caption{\textbf{Comparison of various scheduling techniques.} Using cosine strategy for both produces oversaturation, while using cosine strategy for guidance scheduling and linear for control scheduling produces oversmooth textures at the legs. Results of using both linear scheduling is closest to our strategy, but is lesser textured (notice feet and ears).}
    \label{fig:scheduling_compare}
\end{figure}

\begin{figure}[h]
    \centering
    \includegraphics[width=1\linewidth]{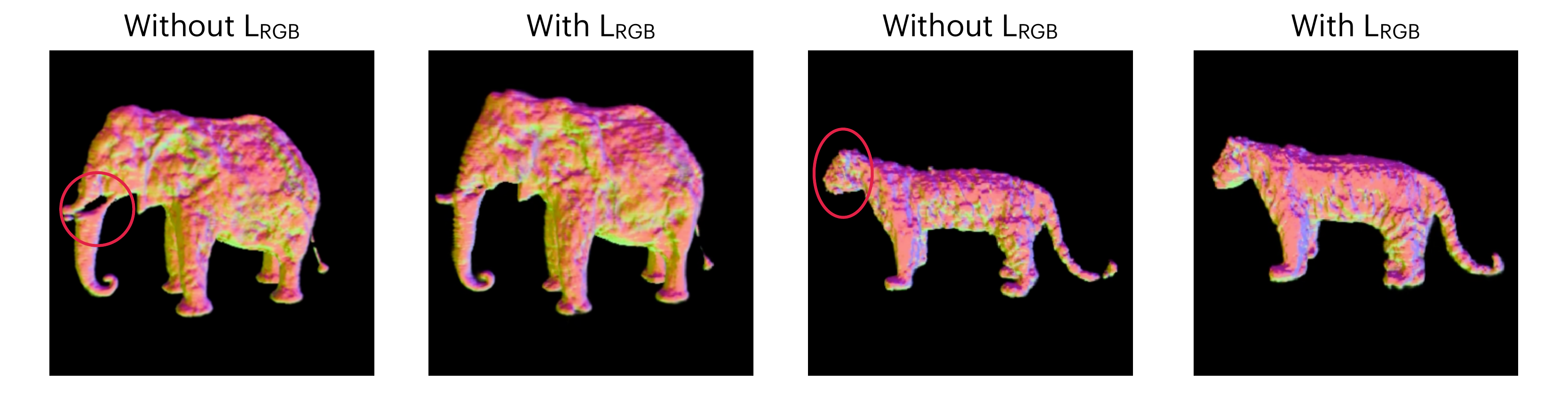}
    \caption{\textbf{Effect of using $\mathcal{L}_{RGB}$.} Not using $\mathcal{L}_{RGB}$ results is hollow geometry and flickering. The chin of the tiger appears and disappears based on view, a view where the chin has disappeared has been chosen.}
    \label{fig:lrgb}
\end{figure}

\section{Additional Ablations}
\label{sec:Additional_Ablations_suppl}
In Fig.~\ref{fig:singlevsmulti} we show that using a single LLM agent performs much worse in generating 3D poses compared to our multi-agent setup which includes Finder, Observer and Modifier GPTs. 

Since our ControlNet guided 3D generation pipeline can produce out-of-domain animals well, the question arises if OpenPose ControlNet can be utilized to generate animals. We show in Fig.~\ref{fig:compare_ControlNet} that OpenPose ControlNet produces artificial looking images for animals. We use a ``human on all fours'' image to obtain the pose for OpenPose and generate a similar keypoint orientation for our TetraPose format. Even though the pose is unnatural for animals, with hips and shoulder very close to spine, TetraPose ControlNet produces clean images following the pose.

\begin{figure}[h]
\newpage
\vspace{-5mm}
 \centering
 \includegraphics[width=0.95\columnwidth]{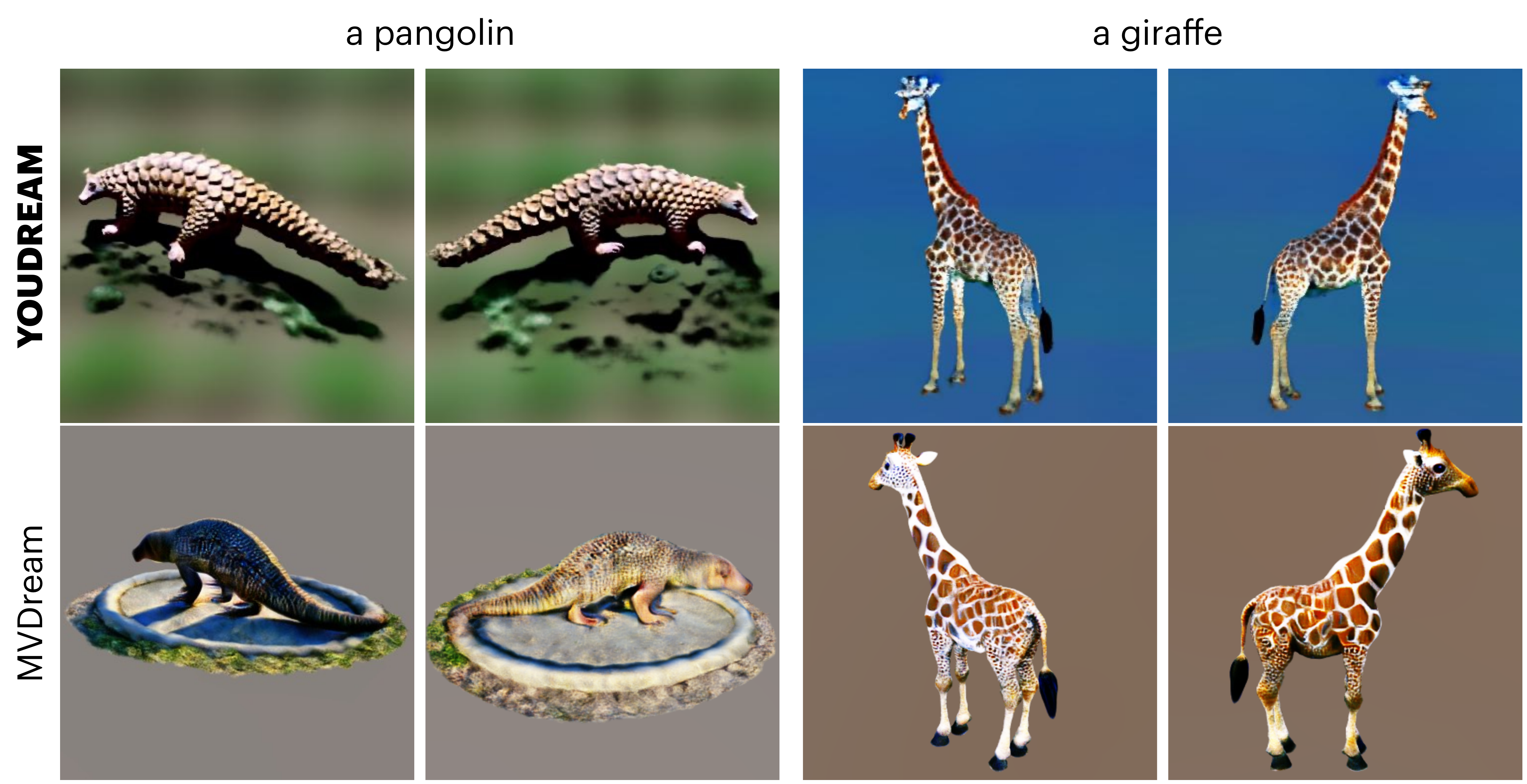}
 \caption{\textbf{More Comparison with MVDream.} We compare our method with MVDream for simple prompts. MVDream results are clearly missing the texture of the scaled body of the pangolin, while their giraffe has a toy-like geometry and hence unnatural. In contrast \textsc{YouDream} produces very realistic results.}
\label{fig:pangolin_vs_giraffe}	
\vspace{-2mm}
\end{figure}

A toy example based on golden ball with wings is presented in Fig.~\ref{fig:ball_compare} to show that text by itself can be ambiguous to convey meaning. When asked for two wings, MVDream produces a modified pair of wings, whereas \textsc{YouDream} follows the user pose control to produce four wings. \textbf{YouDream} performs significantly better for many prompts involving real animals such as pangolin and giraffe. Fig.~\ref{fig:pangolin_vs_giraffe} shows results for both the animals generated using MVDream and \textsc{YouDream}. Even though MVDream is a 3D aware model, it still produces artificial looking results in many cases. While results generated using \textsc{YouDream} are much more natural perceptually and contain realistic textures found in the respective animals.

We show that our method does not require seed tuning for generating consistent results in Fig.~\ref{fig:seed_vary}. Variation in textures and shapes can be seen across seed.

In Fig.~\ref{fig:scheduling_compare} we show the effect of different guidance and control scheduling strategies. Note that for all, guidance scale increases while control scale reduces. 

We show that not using $\mathcal{L}_{RGB}$ loss produces holes and flickering in generated assets. We show the normals for elephant and tiger for this purpose.

\begin{figure}[h]
\newpage
\vspace{-5mm}
 \centering
 \includegraphics[width=0.7\columnwidth]{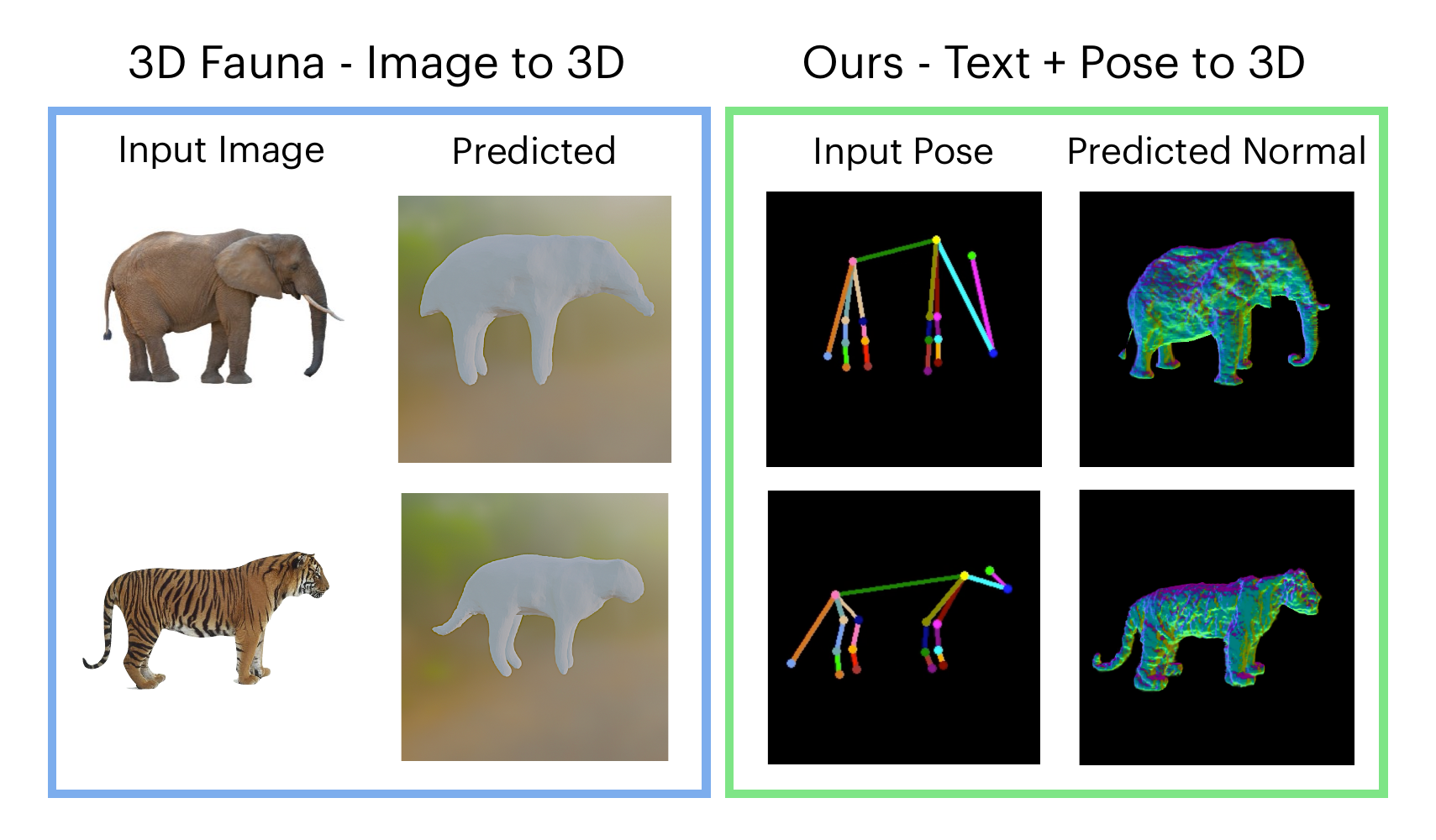}
 \caption{\textbf{Comparison with 3DFauna.} Our method produces more detailed geometry compared to the baseline.}
\label{fig:vs3dfauna}	
\vspace{-2mm}
\end{figure}

\textbf{Comparison with 3D Animal Model.} We compared our method against 3DFauna~\cite{li2024learning}, a 3D animal reconstruction method based on image inputs.  Given an input image 3DFauna failed to capture high-frequency details and follow the input image (see tail and snout in Fig.~\ref{fig:vs3dfauna}), whereas our method produced a highly detailed animal given the input pose and text, which closely followed the input pose control.

\begin{figure}[h]
    \centering
    \includegraphics[width=1\linewidth]{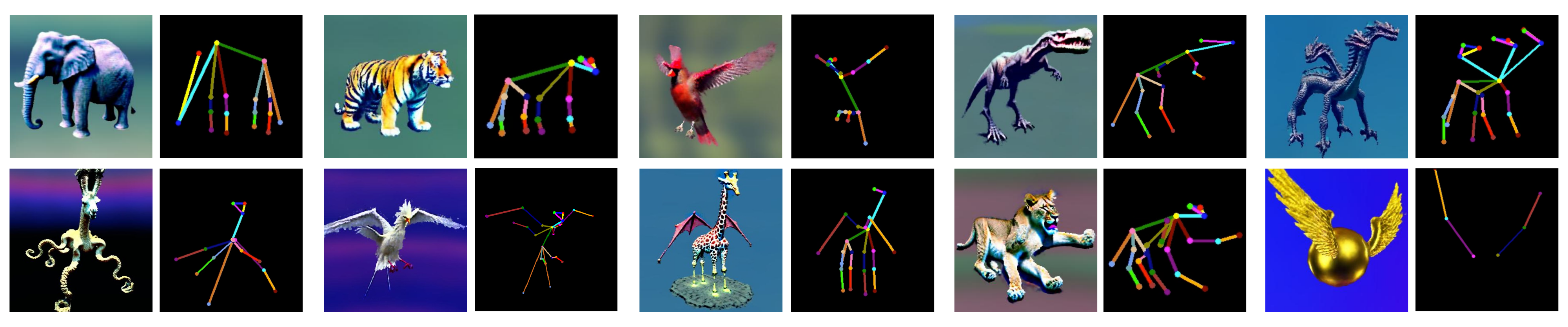}
    \caption{\textbf{Snapshots of 3D poses used for generating objects in the main paper.} For a 2D view of each object, we show the corresponding 2D view of the 3D pose.}
    \label{fig:results_with_pose}
\end{figure}

\section{Implementation details}
\label{sec:implementation_details_suppl}
\textbf{Poses used to generate 3D animals in main paper.} Fig.~\ref{fig:results_with_pose} shows the 2D views of 3D poses used to generate the 3D animals in main paper.

\textbf{TetraPose ControlNet Training:} We used annotated poses from the AwA-pose~\cite{banik2021novel} and Animal Kingdom~\cite{ng2022animal} datasets to train ControlNet in a similar way as the original paper, which uses stable diffusion version 1.5. AwA-pose consists of 10k annotated images covering 35 quadruped animal classes, while Animal Kingdom provides 33k annotated images spanning 850 species, including mammals, reptiles, birds, amphibians, fishes and insects. From a combined set of 43k samples, we carefully selected a subset including only mammals, reptiles, birds, and amphibians. We also eliminated any sample having less than 30\% of its keypoints annotated. The curated dataset consists of 13k annotated samples. To increase diversity in learning, and to improve test-time generation at any scale and transformations, we used a combination of data augmentation strategies consisting of random rotations, translations, and scaling while training so as to handle highly varied and heavily occluded 2D pose samples during 3D generation. The model was trained over 229k iterations with a batch size of 12, a constant learning rate of $1e^{-5}$, on a single Nvidia RTX 6000. The model converged after around 120k iterations and would not overfit even up to 200k iterations, owing in part to the augmentation strategy. 

\begin{figure}[h]
    \centering
    \begin{subfigure}{0.48\columnwidth}
        \centering
        \includegraphics[width=\linewidth]{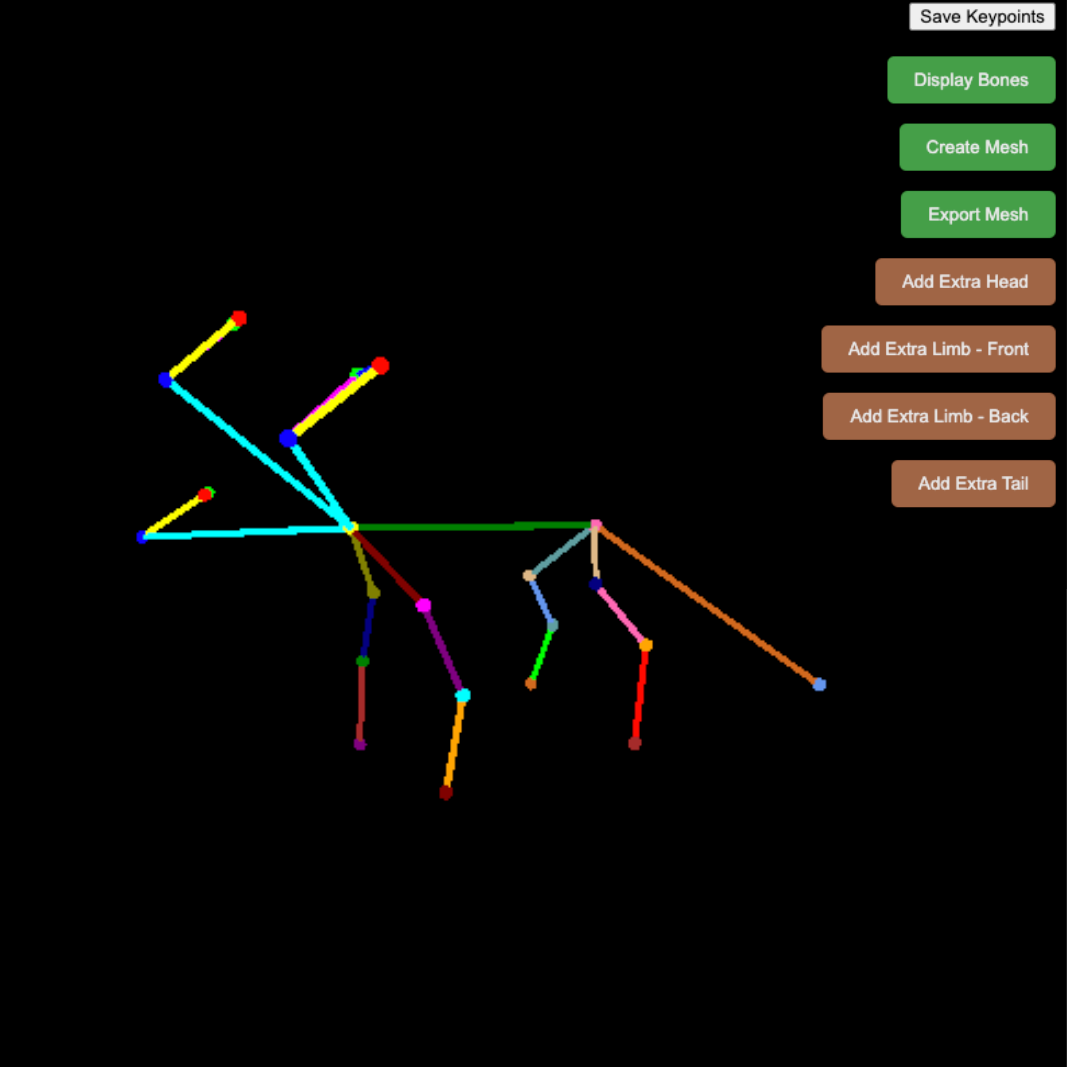}
        \label{fig:pose_editor}
    \end{subfigure}
    \hfill
    \begin{subfigure}{0.48\columnwidth}
        \centering
        \includegraphics[width=\linewidth]{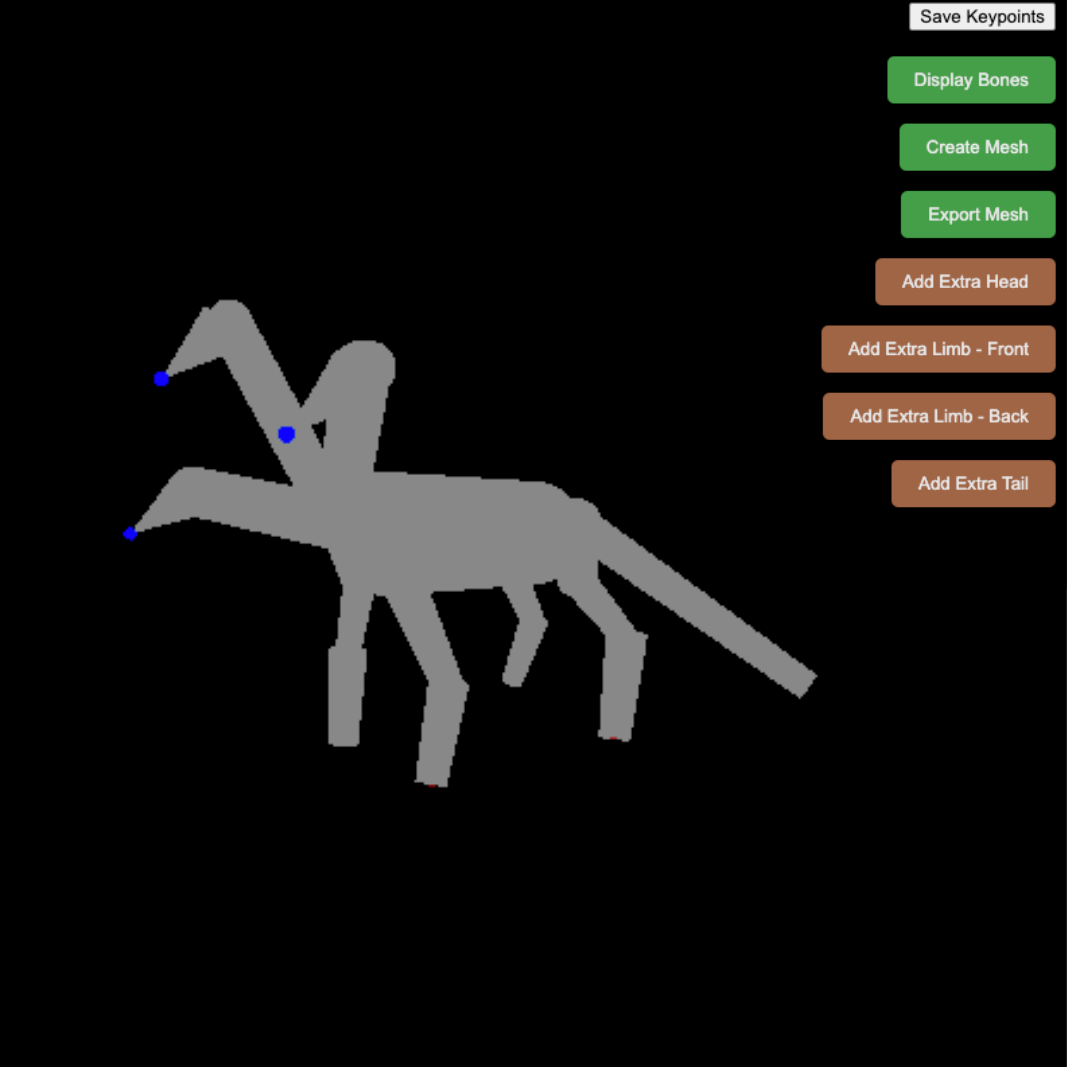}
        \label{fig:mesh_creator}
    \end{subfigure}
    \caption{\textbf{3D Pose editing and Shape generation.} We show snapshots of our 3D pose creator tool with all functionalities.}
    \label{fig:balloon_animal_tool}
\end{figure}

\textbf{3D Pose editing and Shape generation:} We used the following 18 keypoints to represent every quadruped: left eye, right eye, nose, neck end, 4 $\times$ thighs, 4 $\times$ knees, 4 $\times$ paws, back end, and tail end. For the upper limbs of birds, i.e. wings, their front - {thighs, knees, and paws} are defined in accordance with how their upper limbs move. The user can begin with any initial pose from the animal library and modify its keypoints using the Balloon Animal Creator Tool. This tool was developed using THREE.js and can be run on any web browser. The tool provides buttons for the following functions: 1) add extra head. 2) add extra limb - front, 3) add extra limb - back, and 4) add extra tail. After appropriate modification of the pose the user can press the button to \textit{create mesh} around bones. This button press invokes calls to various functions defined to create each body part, based on their natural appearances using simple mesh components such as ellipses (eyes and torso), cylinders (neck, tail, and limbs), and cones (nose). The combined mesh and the corresponding keypoints can be downloaded by clicking the \textit{Export Mesh} and \textit{Save Keypoints} button. An example of this process used for creating the three headed dragon using the Balloon Animal Creator tool is depicted in Fig. \ref{fig:balloon_animal_tool}.

\textbf{Mesh depth guided NeRF initialization}: The mesh downloaded in the previous step was used to provide depth maps to the pre-trained depth guided ControlNet, which produces the gradient loss by SDS, which in turn is used to pre-train the NeRF. The pre-training helps achieve a reasonable initial state for the NeRF weights, which can then be refined in the final pose-guided training stage. The diffusion model was pre-trained for 10,000 iterations using the Adam optimizer with a learning rate of $1e-3$ and a batch size of 1. During training, the camera positions were randomly sampled in spherical coordinates, where the radius, azimuth, and polar angle of camera position were sampled from [1.0, 2.0], [0, 360], and [60, 120]. 

\textbf{Pose-guided SDS for NeRF fine-tuning:} Finally, we fine-tune the NeRF using the pre-trained ControlNet to provide 2D pose guidance to SDS. The gradients computed using the noise residual from SDS were weighted in a similar manner as DreamFusion, where $w(t) = \sigma_{t}^{2}$ and $t$ was annealed using $t = t_{max} - (t_{max} - t_{min})\sqrt{\frac{iter}{total\_iters}}$. We set $t_{max}$ to be 0.98, $t_{min}$ to be $0.4$. Similar to the previous stage, we trained the model over $total\_iters = 10,000$ using the same settings for the optimizer. Using cosine annealing, we reduced the $control_{scale}$ from an initial value of 1 to a final value of 0.2, while updating $guidance_{scale}$ linearly from $guidance_{min} = 50$ to $guidance_{max} = 100$. These settings helped reduce the impact of ControlNet gradually over the training process, while improving quality by gradually increasing strength of classifier-free guidance. The camera positions were randomly sampled as in stage 1, as were the radius, azimuth, and polar angle of the camera. $\lambda_{RGB}$ was set to $0.01$. The 3D avatar representation renders images directly in the RGB space of $\mathbb{R}^{128\times128\times3}$. We use Instant-NGP~\cite{muller2022instant} as the NeRF representation. The pre-training stage, if used, takes less than 12 minutes to complete, while the fine-tuning stage takes less than 40 minutes to complete on a single A100 40GB GPU.  

\textbf{Computational Resources:} All the experiments pertaining to \textsc{YouDream} and 3DFuse were run on Nvidia A100 40GB GPU. Few experiments for MVDream and all experiments of HiFA required running on A100 80GB GPU, while all experiments for Fantasia3D were run on 3xA100 40GB GPUs.

\section{Animal Library}
\label{sec:animal_library_suppl}
Our animal library $\mathcal{B}$ contains a total of 16 animal/pose combinations: 
\begin{itemize}
    \item Giraffe
    \item Elephant
    \item German Shepherd
    \item Eagle - sitting
    \item Eagle - flying
    \item American Crocodile
    \item Tree Frog
    \item Roseate Spoonbill - sitting
    \item Roseate Spoonbill - flying
    \item Raccoon - standing on four legs
    \item Raccoon - standing on two legs
    \item T-Rex
    \item Lizard
    \item Tortoise
    \item Bat
    \item Otter
\end{itemize}
All common animals results shown in this paper are either using these 3D poses or poses modified from one of these by our multi-agent LLM. The library entries are chosen intuitively such that each has significant anatomical variation from the others so as to cover the large range of variety observed in the animal kingdom.

\begin{figure}[h]
    \centering
    \begin{subfigure}{0.98\columnwidth}
        \centering
        \includegraphics[width=\linewidth]{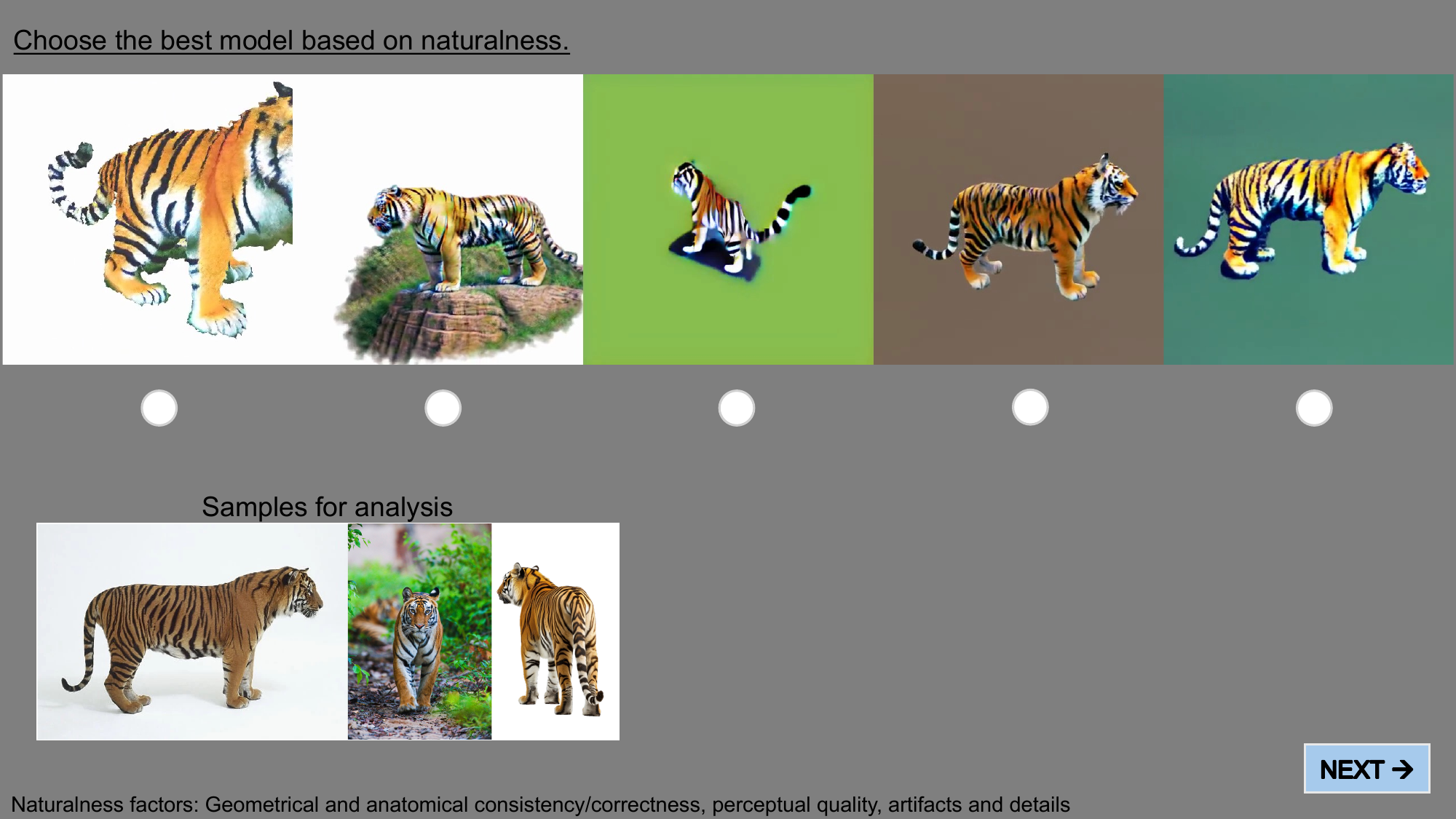}
        \label{fig:interface_naturalness}
    \end{subfigure}
    \caption{\textbf{User Study Interface} for \textit{Naturalness preference}: A snapshot of the interface displaying rotating videos of the results generated by the five chosen models. The user was provided with sample images of the real animal for analyzing anatomical consistency.}
    \label{fig:user_study_interface_1}
\end{figure}

\begin{figure}[h]
    \centering
    \begin{subfigure}{0.98\columnwidth}
        \centering
        \includegraphics[width=\linewidth]{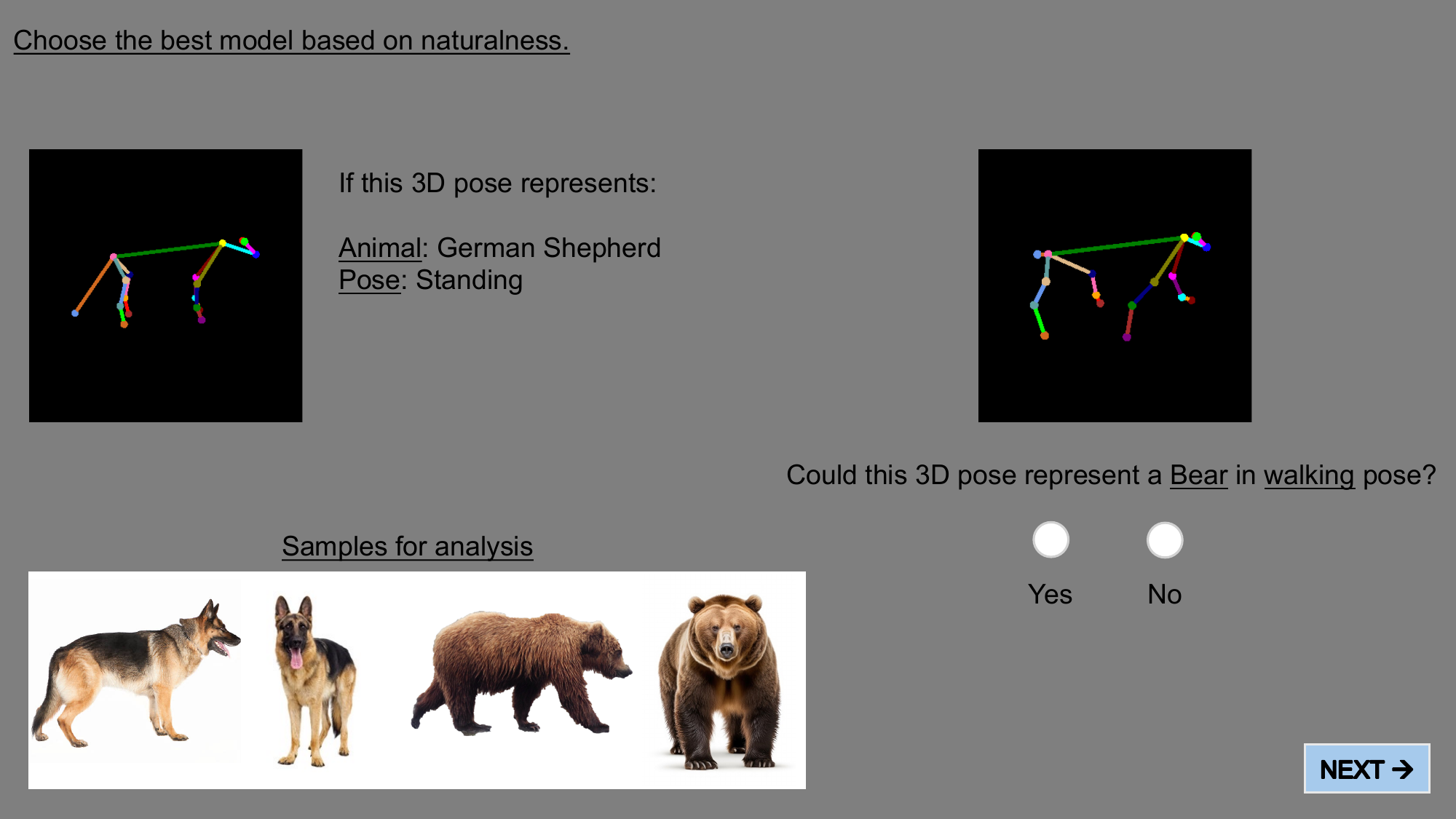}
        \label{fig:interface_T2i}
    \end{subfigure}
    \caption{\textbf{User Study Interface} for \textit{Generated Pose Preference}: A snapshot of the interface displaying rotating pose video of the `reference animal' (left side of the interface) used by the multi-agent LLM for generating a 3D pose of `requested animal' (right side of the interface) in `requested pose'. The user was provided with real samples of the `reference animal' and the `requested animal' (one side-view and one front-view each for better anatomical analysis.}
    \label{fig:user_study_interface_2}
\end{figure}

\section{Multi-agent LLM Implementation Details}
\label{sec:multi-agent_LLM_implementation_details_suppl}
We use the recently released ``GPT-4o'' API of OpenAI with max\_tokens as 4096 and temperature as 0.9. We provide the code for the multi-agent LLM including prompts in Supplementary zip in the file \texttt{GPT\_kpt\_maker.py}

\section{Evaluation using CLIP score}
\label{sec:clip_evaluation_suppl}
Based on the user study, it is clear that users majorly prefer either MVDream or \textsc{YouDream}. Hence we also compute CLIP similarity score for each of the two methods as the average CLIP score over 9 views of each of the 22 prompts used for the user study. Table~\ref{tab:CLIPscore} shows that our method outperforms MVDream based on the CLIP similarity score. We use the ViT-B/32 model for evaluation.

\begin{table}[h]
\centering
\begin{tabular}{l|cc}
           & \textsc{MVDream}                   & \textsc{YouDream}                  \\ \hline
CLIP Score $\uparrow$ & \multicolumn{1}{c}{29.78} & \multicolumn{1}{c}{\textbf{30.86}}
\end{tabular}
\vspace{2mm}
\caption{\textbf{CLIP similarity score} comparison for MVDream and \textsc{YouDream}}
\label{tab:CLIPscore}
\end{table}

\section{User Study Details}
\label{sec:user_study_suppl}
Graduate students at our universities volunteered for participating in the user study. All the information regarding the preferences requested in the study, the judgement criteria, and operating the interface were provided at the beginning of the study. A consent form documenting the purpose of the study, risks involved in the study, duration of the study, compensation details, and contact details for grievances was signed by each user before the beginning of their study session. Screenshots of the user study interfaces are shown in Fig.~\ref{fig:user_study_interface_1} and Fig.~\ref{fig:user_study_interface_2}. The 22 prompts used for generating the 3D assets used in the user study are as follows:
\begin{enumerate}
    \item A giraffe
    \item A lizard
    \item A raccoon standing on two legs
    \item A tiger
    \item A lion
    \item A red male northern cardinal flying with wings spread out
    \item A roseate spoonbill flying with wings spread out
    \item A Tyrannosaurus rex
    \item A pangolin
    \item A bear walking
    \item A horse
    \item A mastiff
    \item A soft cute tiger plush toy, in standing position
    \item An elephant standing on concrete
    \item A dragon with three heads separating from the neck
    \item A realistic mythical bird with two pairs of wings and two long thin lion-like tails
    \item Golden ball with wings
    \item A six legged lioness, fierce beast, pouncing, ultra realistic, 4k
    \item A giraffe with dragon wings
    \item A zoomed out photo of a llama with octopus tentacles body
    \item A zoomed out DSLR photo of a gold eagle statue
    \item Golden ball with two pairs of wings
\end{enumerate}

\begin{figure}[h]
    \centering
    \begin{subfigure}{0.98\columnwidth}
        \centering
        \includegraphics[width=\linewidth]{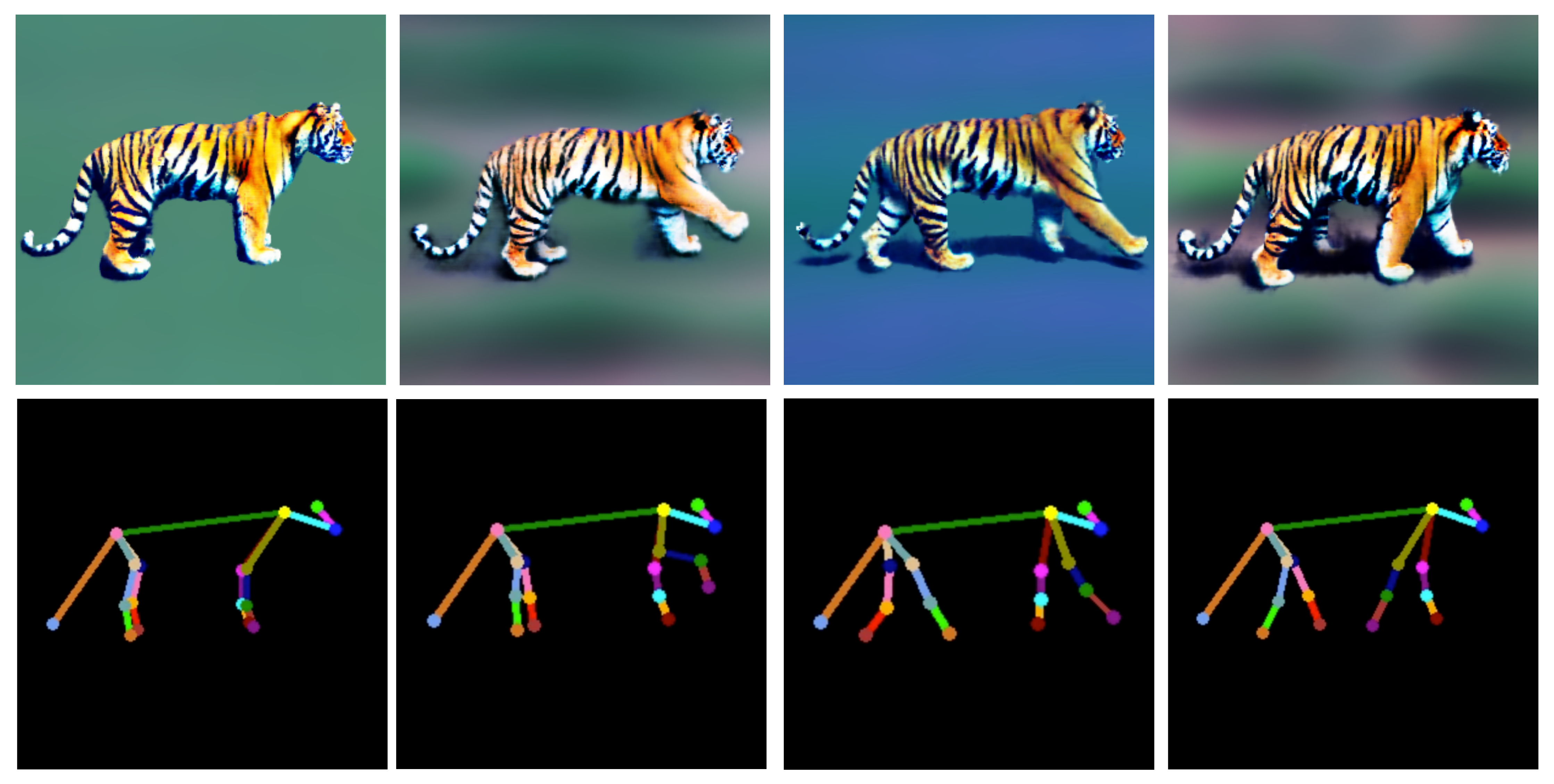}
    \end{subfigure}
    \caption{\textbf{Animation.} Bottom-row: Sampled pose frames from a pose sequence of a tiger walking. Top-row: Camera captured image of the 3D mesh corresponding to the view of the 3D pose shown below it in the bottom-row.}
    \label{fig:animation}
\end{figure}

\section{Animation using Pose Sequence}
\label{sec:animation_suppl}
\textsc{YouDream} can also be used to generate animated videos by generating 3D assets for every pose from a pose sequence. In Fig.~\ref{fig:animation} we show frames chosen from a pose sequence and the corresponding render of their generated 3D mesh. Despite this, generating a longer animation sequence using \textsc{YouDream} would be a highly resource expensive and time consuming task. We hope this work will inspire further exploration of efficient methods for controlled animation.

\section{Limitations and Discussion}
\label{sec:limitations_suppl}
While our method produces high quality anatomically consistent animals, the sharpness and textures can be improved by utilizing a number of tricks used by recent papers. We use a 128$\times$128 NeRF, while our baseline HiFA uses 512$\times$512, while MVDream uses 256$\times$256. We use a smaller NeRF for the sake of lower time complexity compared to baselines. Other tactics such as using DMTet or regularization techniques are also plug-and-play for our method and may improve sharpness.

We show several diverse examples of automatically generating common animals found in nature. However there could exist unusually shaped animals whose 3D pose cannot be satisfactorily generated using our multi-agent LLM setup. In these cases, manual editing of 3D pose might be required over the LLM generated 3D pose. However, we believe our pose editor tool is highly interactive and user-friendly, thus requires very low human effort to modify poses.

\textbf{Broader Impact.} AI generated art has been widely used in recent times. \textsc{YouDream} enables artists to gain more control over their creations, thus making the process of content creation easier. As our method uses Stable Diffusion, it inherits the biases of that model. TetraPose ControlNet training uses existing open-source animal pose datasets instead of internet scraped images, hence avoiding any copyright issues.

\newpage
\textbf{Licenses}
\begin{table}[h]
\centering
\resizebox{\columnwidth}{!}{%
\begin{tabular}{lll}
\hline
URL & Citation & License \\
\hline
\url{https://github.com/JunzheJosephZhu/HiFA} & ~\cite{zhu2023hifa} & Apache License 2.0 \\
\url{https://github.com/KU-CVLAB/3DFuse} & ~\cite{seo2023let} & N/A \\
\url{https://github.com/Gorilla-Lab-SCUT/Fantasia3D} & ~\cite{chen2023fantasia3d} & Apache License 2.0 \\
\url{https://github.com/bytedance/MVDream} & ~\cite{shi2023mvdream} & MIT License \\
\url{https://github.com/lllyasviel/ControlNet} & ~\cite{zhang2023adding} & Apache License 2.0 \\
\url{https://github.com/prinik/AwA-Pose} & ~\cite{banik2021novel} & MIT License \\
\url{https://github.com/sutdcv/Animal-Kingdom} & ~\cite{xu2023animal3d} & N/A \\
\hline
\end{tabular}}
\end{table}

\end{document}